\documentclass[final,3p,number,sort&compress]{elsarticle}
\usepackage{graphicx}
\usepackage{amssymb}
\usepackage{amsmath}
\usepackage{tensor}
\usepackage{amsmath}
\usepackage{hyperref}
\usepackage{xcolor}
\usepackage{mathtools}
\usepackage{floatrow}

\usepackage{float}
\usepackage{array}
\usepackage{booktabs}
\usepackage{lineno}
\usepackage[numbers]{natbib}
\floatstyle{plaintop}
\restylefloat{table}

\usepackage{subfig}

\newlength\tbspace
\newcolumntype{L}{l<{\hspace{\tbspace}}}

\usepackage{array}
\usepackage{tabulary}
\newcolumntype{K}[1]{>{\centering\arraybackslash}m{#1}}

\usepackage{tikz}
\newcommand{\dittotikz}{%
\tikz{
\draw [line width=0.12ex] (-0.2ex,0) -- +(0,0.8ex)
(0.2ex,0) -- +(0,0.8ex);
\draw [line width=0.08ex] (-0.6ex,0.4ex) -- +(-1.5em,0)
(0.6ex,0.4ex) -- +(1.5em,0);
}%
}

\makeatletter
\def\ps@pprintTitle{%
\let\@oddhead\@empty
\let\@evenhead\@empty
\def\@oddfoot{}%
\let\@evenfoot\@oddfoot}
\makeatother

\usepackage{amssymb} 
\newcommand{\cf} {\boldsymbol{\mathfrak{f}}}

\newcommand{\cx}{\boldsymbol{\raisebox{1.5pt}{$\chi$}}}

\newcommand{\dtheta}{\boldsymbol{\dot{\theta}}}
\newcommand{\ddtheta}{\boldsymbol{\ddot{\theta}}}
\newcommand{\dbomega}{\boldsymbol{\dot{\omega}}}

\newcommand{\btimes}{\boldsymbol{\times}}

\newcommand{\dbe}{\boldsymbol{\dot{e}}}
\newcommand{\dbE}{\boldsymbol{\dot{E}}}
\newcommand{\dbR}{\boldsymbol{\dot{R}}}
\newcommand{\dbM}{\boldsymbol{\dot{M}}}

\newcommand{\dbp}{\boldsymbol{\dot{p}}}

\newcommand{\BI}{\boldsymbol{\Bar{I}}}

\newcommand{\dTbomega}{\boldsymbol{\Tilde{\dot \omega}}}
\newcommand{\Tbomega}{\boldsymbol{\Tilde{\omega}}}

\def\R{\mathbb{ R}}

\newcommand{\mJ}{\mathcal J}

\newcommand{\mbJ}{\boldsymbol \mJ}

\newcommand{\bB}{\boldsymbol B}
\newcommand{\bC}{\boldsymbol C}

\newcommand{\bE}{\boldsymbol E}

\newcommand{\bI}{\boldsymbol I}

\newcommand{\bJ}{\boldsymbol J}
\newcommand{\bdJ}{\boldsymbol{\dot{J}}}

\newcommand{\bK}{\boldsymbol K}

\newcommand{\bM}{\boldsymbol M}

\newcommand{\bR}{\boldsymbol R}
\newcommand{\bS}{\boldsymbol S}

\newcommand{\bW}{\boldsymbol W}

\newcommand{\bY}{\boldsymbol Y}

\newcommand{\ba}{\boldsymbol a}

\newcommand{\be}{\boldsymbol e}
\newcommand{\vf}{\boldsymbol f}
\newcommand{\bg}{\boldsymbol g}

\newcommand{\bn}{\boldsymbol n}

\newcommand{\bp}{\boldsymbol p}
\newcommand{\bq}{\boldsymbol q}

\newcommand{\bs}{\boldsymbol s}

\newcommand{\bx}{\boldsymbol x}

\newcommand{\hba}{\boldsymbol{\hat{a}}}
\newcommand{\hbb}{\boldsymbol{\hat{b}}}
\newcommand{\hbc}{\boldsymbol{\hat{c}}}
\newcommand{\hbd}{\boldsymbol{\hat{d}}}

\newcommand{\hbu}{\boldsymbol{\hat{u}}}
\newcommand{\hbv}{\boldsymbol{\hat{v}}}
\newcommand{\hbw}{\boldsymbol{\hat{w}}}

\newcommand{\dhbb}{\boldsymbol{\dot{\hat{b}}}}
\newcommand{\dhbc}{\boldsymbol{\dot{\hat{c}}}}

\newcommand{\dhbu}{\boldsymbol{\dot{\hat{u}}}}
\newcommand{\dhbv}{\boldsymbol{\dot{\hat{v}}}}
\newcommand{\dhbw}{\boldsymbol{\dot{\hat{w}}}}

\newcommand{\bpi}{\boldsymbol \pi}

\newcommand{\btheta}{\boldsymbol \theta}

\newcommand{\etab}{\boldsymbol \eta}

\newcommand{\brho}{\boldsymbol \rho}
\newcommand{\btau}{\boldsymbol \tau}
\newcommand{\bomega}{\boldsymbol \omega}

\newcommand{\blambda}{\boldsymbol \lambda}
\newcommand{\bLambda}{\boldsymbol \Lambda}

\newcommand{\psB}{\bB^\dag}

\begin{document}

\begin{frontmatter}
\title{Dynamic Models of Spherical Parallel Robots for \\ ~Model-Based Control Schemes}
\author[ARAS]{Ali Hassani}
\ead{hassani@email.kntu.ac.ir}
\author[ARAS]{Abbas Bataleblu}
\ead{a.bataleblu@mail.kntu.ac.ir}
\author[ARAS]{S. A. Khalilpour}
\ead{khalilpour@ee.kntu.ac.ir}
\author[ARAS]{Hamid D. Taghirad\corref{cor1}}
\ead{taghirad@kntu.ac.ir}
\author[LAVAL]{Philippe Cardou}
\ead{pcardou@gmc.ulaval.ca}
\cortext[cor1]{Corresponding author}
\address[ARAS]{Advanced Robotics and Automated Systems (ARAS), Faculty of Electrical and Computer Engineering,\\
K.N. Toosi University of Technology, Tehran, Iran.}
\address[LAVAL]{Department of Mechanical Engineering, Robotics Laboratory, Laval University, Quebec City, QC G1V 0A6, Canada.}

\begin{abstract}
\textcolor{black}{In this paper, derivation of different forms of dynamic formulation of spherical parallel robots (SPRs) is
investigated. These formulations include the explicit dynamic forms,
linear regressor, and Slotine-Li (S-L) regressor, which are required
for the design and implementation of the vast majority of
model-based controllers and dynamic parameters identification
schemes.} \textcolor{black}{To this end}, the implicit dynamic of SPRs is first formulated \textcolor{black}{using the principle of
virtual work in task-space}, and then by using an extension, their
explicit dynamic formulation is derived. \textcolor{black}{The dynamic equation is then analytically reformulated into linear and
S-L regression form with respect to the inertial parameters, and by
using the Gauss-Jordan procedure, it is reduced to a unique and
closed-form structure.} Finally, to illustrate the effectiveness of
the proposed method, two different SPRs, namely, the ARAS-Diamond,
and the 3-\underline{R}RR, are examined as the case studies. The
obtained results are verified by using the
MSC-ADAMS\textsuperscript{\textregistered} software,
\textcolor{black}{and are shared to interested audience for public
access.}
\end{abstract}

\begin{keyword}
Spherical Parallel Robots 
\sep Virtual Work Method
\sep Explicit Dynamic Formulation
\sep Spherical Parallel Robots Dynamics
\sep Model-Based Control
\end{keyword}
\end{frontmatter}

\section{Introduction}
\label{intro} The demand for precise robotic manipulators is
consistently increasing in the industry. Parallel robots
\textcolor{black}{(PRs)} may suitably address this requirement.
Having closed kinematic chains in their structure, PRs often possess
higher stiffnesses, accuracies, speeds, and accelerations than their
serial counterparts. Spherical parallel robots
(\textcolor{black}{SPRs}) are a special category of PRs, in which
the moving platform \textcolor{black}{and the other moving links},
are constrained to rotate about a single point, namely, the center
of rotation (CR) \textcolor{black}{\cite{bai2019review}}
\footnote{\textcolor{black}{It should be noted that in a group of
PRs, the moving-platform and some of the links are constrained to
rotate about CR, for example 3-\underline{U}PU wrist. These robots
are outside the scope of this article.}}. Although SPRs are made in
various designs and target different applications, perhaps their
substantial utilization is to rotate a specific object about a
specific point, such as camera attitude \cite{li2015design} or
minimally invasive surgeries (MIS) \cite{bataleblu2016robust}.

\textcolor{black}{The dynamic models of robots are needed for
model-based controller design, dynamic performance analysis, robot
design, and system identification. However, deriving the dynamic
model of PRs is a challenging issue, due to inherent complexity, due
to their closed-loop structure and kinematic constraints. The issue
of the dynamic derivation of PRs is a popular topic in the literature
and has been addressed in various papers
\cite{Angeles_1988,Khalil_1995,Codourey_1998,Dasgupta_1998,wang1998new,Tsai_1999,Khalil_2004,Khalil_2007,
Abdellatif_2009,D_az_Rodr_guez_2010,Oftadeh_2010,Park_2018,M_ller_2019}
and text-books
\cite{tsai1999robot,angeles2002fundamentals,shah2013dynamics,taghirad2013parallel,briot2015dynamics,Staicu_2019}.
In general, the dynamic models of PRs may be derived in the joint-
or task-space coordinates. However, it is advantageous to express
the actuating joint generalized forces of PRs as a function of the
task space variables, since the natural description of PRs dynamics
is in the task space, and in addition, the variables to be
controlled are naturally defined in the task space
\cite{Dasgupta_1998,Paccot_2009}.}

\textcolor{black}{There are several different forms of robot
dynamics that are suitable for designing model-based controllers and
dynamic calibration procedures. The most common form is the explicit
dynamics form, in which robot dynamics is divided into three
components: mass terms, Coriolis and centrifugal acceleration terms,
and gravity terms. The common inverse dynamic controller (IDC),
which is the basis of many advanced model-based controllers, is
based on this form. However, there are always structural and
parametric uncertainties in the robot dynamic model, which can
reduce the performance of IDC \cite{taghirad2013parallel}, and that
makes IDC to be the basis for further robust and adaptive
controllers for motion tracking control of PRs.}

\textcolor{black}{The second derivation form of robot dynamics is
the linear regression form. A large class of adaptive control
schemes~\cite{craig1987adaptive,spong1990adaptive}, as well as many
dynamic calibration procedures \cite{grotjahn2004identification},
requires a regression form that is linear with respect to the
parameters of the robot. Thanks to the linear form of the robot
dynamics, dynamic model uncertainties may be estimated by adaptive
controllers in the feedback loop, or identified using linear
regression techniques \cite{wu2010overview}. Furthermore, there is
another type of linear regressor form of the robot dynamics
introduced by Slotine and Li in~\cite{slotine1987adaptive}. This
regressor form is more suitable for the implementation of adaptive
controller structures. In Slotine-Li (S-L) regressor, restrictive
requirements are eliminated, such as the need for the acceleration
measurement and the computation of the inverse mass matrix.}

\textcolor{black}{The main objective of this paper is to derive
different forms of dynamics of a general SPR, in order to design
different model-based controllers and dynamic identification
schemes. By examining the literature on the subject of various
controllers designed for SPRs, reported in \autoref{Table:List}, it
can be seen that most of reported structures used kinematic-based or
simple model-based controllers~\cite{bai2019review}. Furthermore, it
is quite clear that the controllers that use the dynamic information
in their control law, may lead to better performance
\cite{Paccot_2009,taghirad2013parallel}}.
\begin{table}[hb!]
\caption{Review of notable control schemes for motion control of SPRs.}\label{Table:List}
\begin{center}
\begin{tabular}{ |K{2.4cm}|K{0.8cm}|K{1.3cm}|K{1.4cm}|K{2.6cm}| }
\hline \scriptsize{\textbf{Controller}}&
\scriptsize{\textbf{Ref}}&\scriptsize{\textbf{Method}}
&\scriptsize{\textbf{Robot}} &\scriptsize{\textbf{Considerations}} \\
\hline
\scriptsize{PID} & \scriptsize{\cite{malosio2012spherical}} & \scriptsize{Practical} & \scriptsize{PKAnkle} & \scriptsize{Poor performance} \\ \hline
\scriptsize{PID/Kinematic sliding mode} & \scriptsize{\cite{Hesar_2014}} & \scriptsize{Practical} & \scriptsize{2-DOF 5R} & \scriptsize{\dittotikz} \\ \hline
\scriptsize{PD/PID} & \scriptsize{\cite{Safaryazdi_2017}} & \scriptsize{Practical} & \scriptsize{2-DOF 5R} & \scriptsize{Only Stabilization} \\ \hline
\scriptsize{Kinematic sliding mode} & \scriptsize{\cite{Danaei_2017}} & \scriptsize{Practical} & \scriptsize{2-DOF 5R} & \scriptsize{\dittotikz} \\ \hline
\scriptsize{Kinematic robust adaptive} & \scriptsize{\cite{Rad_2020}} & \scriptsize{Practical} & \scriptsize{3-\underline{R}RR} & \scriptsize{\dittotikz} \\ \hline
\scriptsize{$\mathcal H_\infty$} & \scriptsize{\cite{Abbas_RJ}} &
\scriptsize{Practical} & \scriptsize{ARAS-Diamond} &
\scriptsize{Model Identification} \\
\specialrule{.1em}{.05em}{.05em}
\scriptsize{PD+G} & \scriptsize{\cite{Saafi_2014}} & \scriptsize{Simulation} & \scriptsize{Redundant 3-RRR} & \scriptsize{Uncertainties in the dynamic model} \\ \hline
\scriptsize{PD+G} & \scriptsize{\cite{Birglen_2002}} & \scriptsize{Practical} & \scriptsize{SHaDe} & \scriptsize{\dittotikz} \\ \hline
\scriptsize{PD/IDC}& \scriptsize{\cite{Liu_2020}} & \scriptsize{Simulation} & \scriptsize{3-\underline{R}RR} & \scriptsize{\dittotikz} \\ \hline
\scriptsize{IDC}& \scriptsize{\cite{Li_2018}} & \scriptsize{Simulation} & \scriptsize{3-\underline{R}RR} & \scriptsize{\dittotikz} \\ \hline
\scriptsize{Dynamic robust adaptive switching learning}& \scriptsize{\cite{Li_2019}} & \scriptsize{Simulation} & \scriptsize{3-\underline{R}RR} & \scriptsize{\dittotikz} \\ \hline
\end{tabular}
\end{center}
\end{table}

\textcolor{black}{It seems that a lack of sufficient knowledge about
derivation of dynamic model forms of SPRs has limited the use of
model-based controllers for SPRs. In order to investigate the
different methods for deriving the dynamic models of different SPRs,
we re-examined the literature, and report the resuly in
\autoref{Table:List2}.}
\begin{table}[ht!]
\caption{Review of notable dynamic models for different SPRs.}\label{Table:List2}
\begin{center}
\begin{tabular}{ |c|K{1.8cm}|K{1.4cm}|K{1.4cm}|K{1.4cm}|K{2cm}| }
\hline
\scriptsize{\textbf{Ref}}& \scriptsize{\textbf{Approach}}&\scriptsize{\textbf{Explicit Dynamics}} &\scriptsize{\textbf{Linear Regressor}} &\scriptsize{\textbf{S-L Regressor}}&\scriptsize{\textbf{Case Study}} \\ \hline
\scriptsize{\cite{Arian_2016}} & \scriptsize{Newton-Euler} &
\scriptsize{$\btimes$} & \scriptsize{$\btimes$} &
\scriptsize{$\btimes$} & \scriptsize{2DOF-5R} \\ \hline
\scriptsize{\cite{Wu_2014}} & \scriptsize{Lagrange} &
\scriptsize{$\btimes$} & \scriptsize{$\btimes$} &
\scriptsize{$\btimes$} & \scriptsize{3-\underline{R}RR} \\ \hline
\scriptsize{\cite{Gallardo_2003,Ruggiu_2010}} & \scriptsize{Virtual
Work} & \scriptsize{$\btimes$} & \scriptsize{$\btimes$} &
\scriptsize{$\btimes$} & \scriptsize{2DOF-5R} \\ \hline
\scriptsize{\cite{Enferadi_2010,Enferadi_2018}} &
\scriptsize{\dittotikz} & \scriptsize{$\btimes$} &
\scriptsize{$\btimes$} & \scriptsize{$\btimes$} &
\scriptsize{3-\underline{R}RP} \\ \hline
\scriptsize{\cite{Staicu_2009}} & \scriptsize{\dittotikz} &
\scriptsize{$\btimes$} & \scriptsize{$\btimes$} &
\scriptsize{$\btimes$} & \scriptsize{3-\underline{R}RR} \\
\specialrule{.1em}{.05em}{.05em}
\scriptsize{\cite{abedloo2014closed}} & \scriptsize{Gibbs-Appelle}
& \scriptsize{\checkmark} & \scriptsize{$\btimes$} &
\scriptsize{$\btimes$}& \scriptsize{3-\underline{R}RR} \\ \hline
\scriptsize{\cite{Akbarzadeh_2012}} & \scriptsize{Natural
Orthogonal Complement} & \scriptsize{\checkmark} &
\scriptsize{$\btimes$} & \scriptsize{$\btimes$}&
\scriptsize{3-\underline{R}RP} \\ \hline
\scriptsize{\cite{Di_Gregorio_2003,Li_2018,Li_2019}} &
\scriptsize{Lagrange} & \scriptsize{\checkmark} &
\scriptsize{$\btimes$} & \scriptsize{$\btimes$} &
\scriptsize{3-\underline{R}RR} \\ \hline
\scriptsize{\cite{Akbarzadeh_2010,Zarkandi_2021}} &
\scriptsize{Virtual Work} & \scriptsize{\checkmark} &
\scriptsize{$\btimes$} & \scriptsize{$\btimes$} &
\scriptsize{3-\underline{R}RP, 3-P\underline{R}R} \\
\specialrule{.1em}{.05em}{.05em}
\scriptsize{\cite{danaei2017dynamic}} & \scriptsize{Virtual Work} &
\scriptsize{$\btimes$} & \scriptsize{\checkmark} &
\scriptsize{$\btimes$} & \scriptsize{2DOF-5R} \\
\specialrule{.1em}{.05em}{.05em}
\scriptsize{This Paper} & \scriptsize{Virtual Work} &
\scriptsize{\checkmark} & \scriptsize{\checkmark} &
\scriptsize{\checkmark} & \scriptsize{2DOF-5R, and
3-\underline{R}RR} \\ \hline
\end{tabular}
\end{center}
\end{table}

As it is seen in this table in most cases, the derived dynamic
models are not complete enough to be used in the design of
model-based controllers. Deriving the appropriate dynamic model to
identify the robot's dynamic parameters is more critical, where
Ref.~\cite{danaei2017dynamic} addresses this issue, but only for
the special case of 2DOF-5R SPRs. In addition, the derivation of the
S-L regressor, which is necessary for designing various adaptive
controllers, is not reported for any SPR robot in the literature.

In this paper, different forms of the dynamics of SPRs will be
examined in detail using the principle of virtual work while the
foremost contributions are summarized as follows:
\begin{enumerate}
\item \textcolor{black}{Providing a systematic method of deriving explicit dynamic model of a general SPR.}

\item \textcolor{black}{Expanding the implicit dynamic formulation to derive linear regressor form,
which allows the design a class of adaptive controllers
\cite{craig1987adaptive,spong1990adaptive}}.

\item \textcolor{black}{Deriving the reduced regressor matrix using the unique solution,
in order to use a wide range of robot dynamic calibration schemes,
for example, BIRDy (Benchmark for identification of robot dynamics) MATLAB toolbox \cite{Leboutet_2021}.} It should be noted that the BIRDy toolbox can only be used for serial manipulators.

\item \textcolor{black}{Using the explicit dynamic formulation and linear regressor form
to derive the S-L regressor form of SPRs, in order to be used in
many adaptive position/force control schemes, as in
\cite{spong2008robot,harandi2021adaptive}.}

\end{enumerate}

\textcolor{black}{To the best of the authors' knowledge, such formulation has never been presented in the literature before.
Previous formulations usually are case dependent, which cannot be
generalized into a systematic method of dynamic formulation. Or they
provide only one of these dynamic forms, which is not sufficient to
be used for the advanced control structures of the position/force of
SPRs.} \textcolor{black}{To demonstrate the scope of the dynamic formulation derived in this paper in terms of structure synthesis,
different SPRs are referred to, while two different SPRs, namely,
ARAS-Diamond (2-DOF 5R), and 3-\underline{R}RR, are considered as
the case studies.} \textcolor{black}{The results of the derived
dynamics model are made publicly available
\footnote{\url{Github.com/aras-labs/SPRs\_Dynamic\_Model}}.}

\textcolor{black}{The remainder of this paper is organized as
follows. Implicit and explicit dynamic analysis are presented in
section 2, while section 3 describes the linear regressor and the
Slotine-Li regressor analysis of SPRs. Then, the dynamic analysis on
two case studies is reported in section 4, and the simulation and
validation results by using
MSC-ADAMS\textsuperscript{\textregistered} software is given in
section 5. Finally, the concluding remarks are given in the last
section.}

\section{Dynamic Analysis}
\subsection{Preliminary Definitions}

\begin{figure}[!ht]
\includegraphics[scale=0.7]{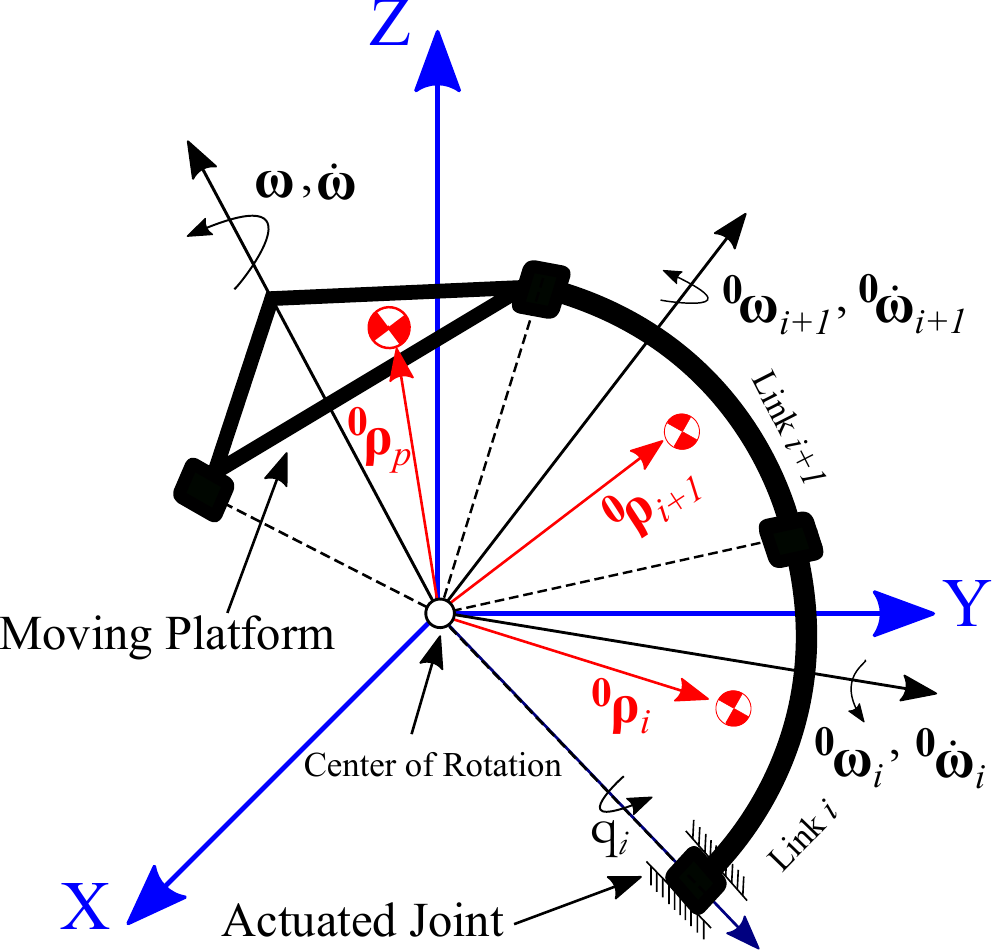}
\caption{View of a SPR} \label{fig:General_Sch}
\end{figure}

\textcolor{black}{In this section, preliminary definitions for the
dynamic formulation of SPRs are given.} The structure of these
robots consists of several links and a moving platform with pure
rotational motion about the CR point. The origin of the base
coordinate system is set at the CR point, as shown in the schematic
of \autoref{fig:General_Sch}.

Throughout this paper, the $i$ and $p$ indices refer to each link
and to the moving platform, respectively. Moreover,
$\bq=\begin{bmatrix} q_1 & ... & q_j \end{bmatrix}^T$ denotes the
actuated joint variables, \textcolor{black}{while $\cx$ represents
independent generalized parameters of the moving platform. In SPRs,
the moving platform has pure rotation about the CR point, and
therefore, $\cx$ may be considered as $\cx=\btheta=\begin{bmatrix}
\theta_1, \theta_2, \theta_3 \end{bmatrix}^T$.} In addition, to
describe the orientation of the moving platform in the task space
coordinate, different approaches are reported in the literature. In
this paper, Euler angles and the spherical base coordinate system
are adopted. \textcolor{black}{Besides, $S(\ba)$ denotes the
$3\times 3$ skew-symmetric matrix generated from the vector
$\ba=[a_1,a_2,a_3]^{T}$.}


\subsection{The Principle of Virtual Work for Parallel Robots}\label{Sec:Virt}
The principle of virtual work for a PR with several
links and a moving platform, may be stated as
\cite{Tsai_1999}:
\begin{equation}\label{eq:Virtual_Work}
\delta W= {\delta \bq}^{T} \cdot \btau + {\delta \cx_p}^T \cdot \cf_{p} +
\sum_{i=1}^{k} \left( {\delta \cx_i} ^{T} \cdot \cf_i \right) =0,
\end{equation}
where \textcolor{black}{$\delta \cx_p=\left[\delta{\bx_p}^T,
\delta{\btheta_p}^T \right]^T$ and $\delta
\cx_i=\left[\delta{\bx_i}^T, \delta{\btheta_i}^T \right]^T$
represent the virtual displacements of the CGs of the moving
platform and of link $i$, respectively.} Furthermore, $\cf_p$ and
$\cf_i$ denote the inertial wrenches respectively applied at the
moving platform CG and the \textcolor{black}{$i$th} link CG.
{Moreover, $k$ indicates the number of links.} These inertial
wrenches take the form:
\begin{subequations}\label{eq:Inertial_Forces}
\begin{align}
&\cf_{p}= \begin{bmatrix} \vf_p \\ \bn_p \end{bmatrix}=
-\begin{bmatrix}
m_p\left(^{0}\mathrm{\ba}_{p}-\textcolor{black}{\bg_0}\right)\textcolor{black}{- \vf_{G_d}} \\ \left(^{0}\mathrm{\bI}_{p}~^{0}\mathrm{\dbomega} +~^{0}\mathrm{ \bomega} \times \left(^{0}\mathrm{\bI}_{p} ~^{0}\mathrm{ \bomega} \right ) \right) \textcolor{black}{- \bn_{G_d} }
\end{bmatrix}, \\
&\cf_{i}=\begin{bmatrix} \vf_i \\ \bn_i \end{bmatrix}=-\begin{bmatrix}
m_i\left(^{0}\mathrm{\ba}_{i}-\textcolor{black}{\bg_0}\right) \\ \left(^{0} \mathrm{\bI}_{i}~
^{0}\mathrm{\dbomega}_i +~^{0}\mathrm{ \bomega}_i \times
\left(^{0}\mathrm{\bI}_{i}~^{0}\mathrm{ \bomega}_i\right ) \right)
\end{bmatrix},
\end{align}
\end{subequations}
in which $^{0}\mathrm{\ba}_{p}$ denotes the linear acceleration of
CG of the moving platform, and $^{0}\mathrm{\bI}_{p}$ demotes the
inertia matrix of the moving platform about its CG and expressed in
the base coordinate system. Moreover, $^{0}\mathrm{ \bomega}$ and
$^{0}\mathrm{\dbomega}$ respectively represent the angular velocity
and acceleration of the moving platform. Similar to this notation,
the same is applied to each link's inertial force with index $i$.
\textcolor{black}{In addition,
$\cf_{G_d}=\left[{\vf_{G_d}}^T,{\bn_{G_d}}^T \right]^T$ is an
external disturbance wrench applied on the moving platform CG.}

By choosing the $\delta \cx= \delta \cx_p$ as the independent
generalized virtual displacements of the PR, virtual displacement of
links, $\delta \bq$, may be related to the virtual displacement of
the moving platform, $\delta \cx$, by the manipulator Jacobian
\textcolor{black}{$\bJ=\left[{\bJ_{v}}^T,{\bJ_{\omega}}
^T\right]^T$} as:
\begin{equation} \label{eq:delq}
\delta \bq=\bJ ~\delta \cx.
\end{equation}

Furthermore, the virtual displacement of the CG of link $i$, $\delta \cx_i$, may be related to $\delta \cx$ by a Jacobian
matrix defined for each link and denoted by
\textcolor{black}{$\bJ_i=\left[{\bJ_{v_i}}^T,{\bJ_{\omega_i}}^T\right]^T$} as:
\begin{equation} \label{eq:delx}
\delta \cx_i=\bJ_i~\delta \cx.
\end{equation}

Substituting \eqref{eq:delq} and \eqref{eq:delx} into
equation of \eqref{eq:Virtual_Work} results in:
\begin{equation}
{\delta \cx} ^T\left(\bJ^{T} \btau +\cf_p+ \sum_{i=1}^{k} {\bJ_i}^{T} \left(\cf_i\right)\right)=0.
\end{equation}

Finally, the implicit dynamics formulation of PRs may be
represented as follows:
\begin{equation}\label{eq:impda}
\cf={\bJ}^{T} \btau = - \left(\cf_p + \sum_{i=1}^{k} {\bJ_i}^{T} \left(\cf_i\right) \right).
\end{equation}

\textcolor{black}{In general, $\cf = \left[{\vf}^T,{\bn}^T
\right]^T= \bJ^{T} \btau$ is defined as a mapping of the actuator
forces $\btau$ {into the space of moving platform wrenches.}}
Therefore, the Jacobian matrix is a squared matrix if the robot is
fully parallel with no redundancy in actuators. Accordingly, in
non-singular configurations, actuator forces may uniquely be derived
as $\btau=\bJ^{-T}\cf$.

As mentioned before, this formulation is based on the kinematic
analysis of the CG of each link and of the moving platform. However,
if external wrenches \eqref{eq:Inertial_Forces} are applied to an
arbitrary point $\mathcal{A}$ distinct from the CG, the external
wrenches equation may be rewritten according to this arbitrary
point. It may be proved that the principle of virtual work for any
arbitrary point $\mathcal{A}$ in PRs is derived as
\cite{codourey1997body}:
\begin{equation}\label{eq:Virtual_Work_Point_A}
\delta W= {\delta \bq}^{T} \cdot \btau + {\delta \cx_{\mathcal{A}_p}}^T \cdot {\cf}_{\mathcal{A}_p} +
\sum_{i=1}^{k} \left( {\delta \cx_{\mathcal{A}_i}} ^{T} \cdot {\cf}_{\mathcal{A}_i} \right) =0,
\end{equation}
in which, \textcolor{black}{$\delta \cx_{\mathcal{A}_p}=\left[\delta{\bx_{{\mathcal{A}_p}}}^T,\delta{\btheta_{\mathcal{A}_p}}^T \right]^T$ and $\delta \cx_{\mathcal{A}_i}=\left[\delta{\bx_{\mathcal{A}_i}}^T,\delta{\btheta_{\mathcal{A}_i}}^T \right]^T$} indicate virtual
displacements of any arbitrary point $\mathcal{A}$ of the moving
platform and each link, respectively. Therefore, inertial forces of the
moving platform and each link with respect to this arbitrary point
may be derived as:
\begin{subequations}\label{eq:Inertial_Forces_II}
\begin{align}
& \cf_{\mathcal{A}_p}=\begin{bmatrix} \vf_{\mathcal{A}_p} \\ \bn_{\mathcal{A}_p} \end{bmatrix}
=-\begin{bmatrix}
m_p \left(^{0}\mathrm{\dbomega} \times~^{0}\mathrm{\brho}_{p} +~^{0}\mathrm{\bomega} \times \left(^{0}\mathrm{\bomega} \times~^{0}\mathrm{\brho}_{p} \right)+\left(^{0}\mathrm{\ba}_{\mathcal{A}_p}-\bg_0\right)\right)\textcolor{black}{-\vf_{d}}\\
\left(^{0}\mathrm{\bI}_{\mathcal{A}_p}~^{0}\mathrm{\dbomega}+~^{0}\mathrm{\bomega} \times \left(^{0}\mathrm{\bI}_{\mathcal{A}_p}~^{0}\mathrm{\bomega}\right)+m_p~^{0}\mathrm{\brho}_p \times \left( ^{0}\mathrm{\ba}_{\mathcal{A}_p}-\bg_0 \right)\right)\textcolor{black}{-\bn_{d}}
\end{bmatrix},\\
& \cf_{\mathcal{A}_i}=\begin{bmatrix} \vf_{\mathcal{A}_i} \\ \bn_{\mathcal{A}_i} \end{bmatrix}=-\begin{bmatrix}
m_i \left(^{0}\mathrm{\dbomega}_i \times~^{0}\mathrm{\brho}_{i} +~^{0}\mathrm{\bomega}_i \times \left(^{0}\mathrm{\bomega}_i \times~^{0}\mathrm{\brho}_{i} \right)+\left(^{0}\mathrm{\ba}_{\mathcal{A}_i}-\bg_0\right)\right)\\
\left(^{0}\mathrm{\bI}_{\mathcal{A}_i}~^{0}\mathrm{\dbomega}_i+~^{0}\mathrm{\bomega}_i \times \left(^{0}\mathrm{\bI}_{\mathcal{A}_i}~^{0}\mathrm{\bomega}_i\right)+m_i~^{0}\mathrm{\brho}_i \times \left( ^{0}\mathrm{\ba_{\mathcal{A}}}_i-\bg_0 \right)\right)
\end{bmatrix},
\end{align}
\end{subequations}
in which, $^{0}\mathrm{\ba}_{\mathcal{A}_p}$ denotes the
acceleration of point $\mathcal{A}_p$ in the base coordinate system.
The vector $^{0}\mathrm{\brho_p}$, points from the arbitrary point
$\mathcal{A}_p$ to the origin of the base coordinate system. It must
be noted that equal notation applies to the inertial forces of each
link with the index of $i$. \textcolor{black}{Moreover, we define
$\cf_{d}=\left[{\vf_{d}}^T,{\bn_{d}}^T \right]^T$ as the external
disturbance wrench applied on the moving platform at the arbitrary
point $\mathcal{A}_p$.}

Therefore, PRs' dynamic formulation may be derived analogously to
the process mentioned in equations of \eqref{eq:delq} and
\eqref{eq:delx}. The only difference, however, is the kinematic
analysis, which is based on the arbitrary points $\mathcal{A}_i$ and
$\mathcal{A}_p$, for each link and the moving platform,
respectively. Thus, in the following, this general formulation is
used to derive a simple form of the SPR dynamics.

\subsection{Implicit Dynamic Analysis of Spherical Parallel Robots}

As mentioned earlier, in SPRs, the moving platform undergoes pure
rotations about the CR point. Accordingly, we propose expressing the
dynamic formulations in a coordinate system with its origin at the
CR. Hence, by selecting $\mathcal{A}_i=\mathcal{A}_p$ as a fixed
point at the origin of the base coordinate system, all links and the
moving platform experience pure rotations about the CR point. As a
result, the virtual linear displacement of points of
\textcolor{black}{$\mathcal{A}_i$ and $\mathcal{A}_p$} always
remains zero. This implies that, the substitution of
\textcolor{black}{$\delta \btheta_{\mathcal{A}_p} $ and $\delta
\btheta_{\mathcal{A}_i}$} into Eq.~\eqref{eq:Virtual_Work_Point_A}
leads to $\delta \bq=\bJ_\omega ~\delta \btheta$ and $\delta
\btheta_i=\bJ_{\omega_i}~\delta \btheta$. Therefore, the dynamic
formulation of SPRs may be accomplished as follows, which would be
interpreted as the implicit dynamic formulation of SPRs:
\begin{equation}\label{eq:implicitdy_Final}
{\bJ_\omega}^{T} \btau=\textcolor{black}{\bn+\bn_d=\bn_{p}+\sum \limits_{i=1}^k \bn_{i}}~,~ \in \R^{m},
\end{equation}
in which, \textcolor{black}{$ {\bJ_\omega}^{T} \btau={\bn+\bn_d}$ represents the mapping of the
actuator forces $\btau$ on moving platform moments. }Moreover, $m$ denotes
the number of generalized task space variables, and the $\bn_{p}$ and $\bn_{i}$
are defined as:
\begin{subequations}\label{eq:implicitdy}
\begin{align}
&\bn_{p}=~^{0}\mathrm{\bI}_{\mathcal{A}_p} ~^{0}\mathrm{\dbomega} +S \left(^{0}\mathrm{\bomega}\right)\left(^{0}\mathrm{\bI}_{\mathcal{A}_p} ~^{0}\mathrm{\bomega}\right)-m_i~S\left (^{0}\mathrm{\brho}_p \right) \bg_0 \label{eq:Implicit_Dy_Op}, \\
&\bn_{i}= {\bJ_{\omega_i}}^{T} \left(^{0}\mathrm{\bI}_{\mathcal{A}_i}~^{0}\mathrm{\dbomega}_i+S\left(^{0}\mathrm{\bomega}_i\right) \left(^{0}\mathrm{\bI}_{\mathcal{A}_i}~^{0}\mathrm{\bomega}_i\right)-m_i~S\left (^{0}\mathrm{\brho}_i\right) \bg_0 \right), \label{eq:Implicit_Dy_Oi}
\end{align}
\end{subequations}
where, $S(\ba)$ represents the $3\times 3$ skew-symmetric matrix of
vector \textcolor{black}{$\ba$ and the} following mapping may be
used for deriving $S\left(^{0}\mathrm{\brho}_i\right)$ and,
likewise, $S\left(^{0}\mathrm{\brho}_p\right)$:
\begin{equation}\label{maping}
S\left(^{0}\mathrm{\brho}_i\right)=~^{0}\mathrm{\bR}_i~S\left(^{i}\mathrm{\brho}_i\right)~{^{0}\mathrm{\bR}_i}^T.
\end{equation}
\textcolor{black}{in which, $^{0}\mathrm{\bR}_i$ denotes the
rotation matrix of each link, respect to base coordinate system, and
$^{i}\mathrm{\brho}_i$ is the center of gravity (CG) position of
link $i$ respect to its CG.}

\subsection{Explicit Dynamic Analysis of Spherical Parallel Robots} \label{EXplicit_Analysis}

As stated before, the orientation of the moving platform of SPRs
might be determined by their generalized coordinate, while its
derivatives represent the angular velocity and the angular
acceleration as $\dtheta={^{0}\mathrm{\bomega}}$, and
$\ddtheta={^{0}\mathrm{\dbomega}}$. In addition, the angular
velocity and acceleration of each link may be expressed as:
\begin{subequations}\label{eq:jacjacjac}
\begin{align}
& ^{0}\mathrm{\bomega}_i=\bJ_{\omega_i} \dtheta \label{jacjac_1},\\
&^{0}\mathrm{\dbomega}_i=\bJ_{\omega_i} \ddtheta+ \bdJ_{\omega_i} \dtheta.
\end{align}
\end{subequations}

In order to derive the explicit form of the dynamic formulations
employing the principle of virtual work, substitute
\eqref{eq:jacjacjac} into the implicit dynamic formulation of
\eqref{eq:implicitdy_Final}.
\textcolor{black}{With some manipulations}, the explicit dynamic formulation for SPRs is written as:
\begin{equation}\label{eq:Explicit}
{\bJ_\omega}^{T} \btau=\textcolor{black}{\bn+\bn_d} =\bM(\btheta)\ddtheta+ \bC(\btheta,\dtheta)\dtheta+\bg(\btheta),
\end{equation}
in which,
\begin{subequations}\label{eq:Explicit_Dynamic_Total}
\begin{align}
&\bM(\btheta)={^{0}\mathrm{\bI}_{\mathcal{A}_p}}+\sum \limits_{i=1}^k \left({\bJ_{\omega_i}}^T~^{0}
\mathrm{\bI}_{\mathcal{A}_i}~\bJ_{\omega_i}\right), \label{eq:M(X)} \\
& \bC(\btheta,\dtheta)=
S(\bomega)~^{0}\mathrm{\bI}_{\mathcal{A}_p}
+\sum \limits_{i=1}^k
{\bJ_{\omega_i}}^{T} \left( ^{0}\mathrm{\bI}_{\mathcal{A}_i}~\bdJ_{\omega_i} + S \left(\bJ_{\omega_i} \dtheta \right)
~^{0}\mathrm{\bI}_{\mathcal{A}_i}~\bJ_{\omega_i} \right), \label{eq:C(X)}\\
& \bg(\btheta)=-m_p~S\left (^{0}\mathrm{\brho}_p \right) \bg_0 -m_i\sum \limits_{i=1}^k {\bJ_{\omega_i}}^{T}~ S(^{0}\mathrm{\brho}_i)~\bg_0. \label{eq:G(x)}
\end{align}
\end{subequations}
In order to use the CG position of each link in the body coordinate
system instead of the base coordinate system, we substitute
\eqref{maping} into \eqref{eq:G(x)}, which yields:
\begin{equation}\label{eq:GG(X)}
\bg(\btheta)=-m_i \left(^{0}\mathrm{\bR}_p~S\left(^{p}\mathrm{\brho}_p\right){{^{0}\mathrm{\bR}_p}^{T}}
\right) \bg_0 -m_i \sum \limits_{i=1}^k {\bJ_{\omega_i}}^{T}
\left(^{0}\mathrm{\bR}_i~S\left(^{i}\mathrm{\brho}_i\right){{^{0}\mathrm{\bR}_i}^{T}}
\right)\bg_0.
\end{equation}

The dynamic matrices of the explicit dynamic formulation of SPRs
obtained in equation \eqref{eq:Explicit} satisfy two important
properties, which might be used in the design of model-based
controllers~\cite{spong2008robot}:

\textbf{Property 1}\label{prop1}. Matrix $\bM(\btheta)$ is positive definite
in all configurations.

\textbf{Property 2}\label{prop2}. The matrix
$\left[\dbM(\btheta,\dtheta)-2\bC(\btheta,\dtheta)\right]$ is
skew--symmetric.

The first property may be easily proved, considering the quadratic
structure of its expression in Eq.~\eqref{eq:M(X)}. Proving the
second property is less straight forward. First, the time derivative
of $\bM(\btheta)$ may be computed as:
\begin{equation}\label{eq:Mdot-2C}
\begin{split}
\dbM= & \left(S({^{0}\mathrm{\bomega}})~^{0}\mathrm{\bI}_{\mathcal{A}_p}+~^{0}
\mathrm{\bI}_{\mathcal{A}_p}~{S({^{0}\mathrm{\bomega}})}^T \right)
+ \sum \limits_{i=1}^k {\bdJ_{\omega_{i}}}^T ~^{0}\mathrm{\bI}_{\mathcal{A}_i}~\bJ_{\omega_i} \\ &+{\bJ_{\omega_{i}}}^T \left(S\left(^{0}\mathrm{\bomega}_i \right)~^{0}\mathrm{\bI}_{\mathcal{A}_i}+~^{0}\mathrm{\bI}_{\mathcal{A}_i}~S\left(^{0}\mathrm{\bomega}_i\right)^T \right)\bJ_{\omega_i}+{\bJ_{\omega_{i}}}^T ~^{0}\mathrm{\bI}_{\mathcal{A}_i}~\bdJ_{\omega_i}.
\end{split}
\end{equation}
Therefore, it can be easily verified that:
\begin{equation}\label{eq:Mdot-2C}
\left(\dbM(\btheta,\dtheta)-2 \bC(\btheta,\dtheta)\right)^T+\left(\dbM(\btheta,\dtheta)-2 \bC(\btheta,\dtheta) \right)=0.
\end{equation}

\section{Regressor Analysis}

\subsection{Slotine-Li Regressor}
In the S-L adaptive controller, the control effort in the task-space is
defined as \cite{slotine1987adaptive}:
\begin{equation}\label{eq:Slotine_Control_Effort}
\bM(\btheta)\ddtheta_r+\bC(\btheta,\dtheta)\dtheta_r+\bg(\btheta)-\bK \bs=\bY_{S}(\btheta,\dtheta,\dtheta_r,\ddtheta_r) \bpi-\bK \bs,
\end{equation}
in which, the error is defined as $\be=\btheta- \btheta_d$, whereas
$\btheta_d$ denotes the desired trajectory in the task-space.
Moreover, the reference velocity of $\dtheta_r$ is defined as
$\dtheta_r=\dtheta_d- \bLambda \be$, and $\bs=\dtheta-
\dtheta_r=\dbe+\bLambda \be$ indicates the sliding surface.
Furthermore, $\bK$ and $\bLambda$ represent symmetric positive
definite gains while,
$\bY_{S}(\btheta,\dtheta,\dtheta_r,\ddtheta_r)$ is known as the
Slotine-Li regressor.

In order to derive this regressor for SPRs, $\bM(\btheta)$, $\bC(\btheta,\dtheta)$, and
$\bg(\btheta)$ are substituted into Eq.~\eqref{eq:Slotine_Control_Effort}. By this
means, the following relation is derived:
\begin{subequations}\label{eq:Slotine_Implicit}
\begin{align}
&\bn_S +\textcolor{black}{\bn_d} = \bn_{S_p}+ \sum \limits_{i=1}^k \bn_{S_i},
\\& \bn_{S_p}=\left(^{0}\mathrm{\bI}_{\mathcal{A}_p}\right) {\ddtheta}_r + \left(S\left({^{0}\mathrm{\bomega}}\right)~^{0}\mathrm{\bI}_{\mathcal{A}_p}\right) {\dtheta}_r-m_p~S\left (^{0}\mathrm{\brho}_p \right) \bg_0,\\
\begin{split}
&\bn_{S_i}= {\bJ_{\omega_i}}^{T} \left(\left(^{0}\mathrm{\bI}_{\mathcal{A}_i}~\bJ_{\omega_i} \right) \ddtheta_r+ \left( ^{0}\mathrm{\bI}_{\mathcal{A}_i}~\bdJ_{\omega_i} + S \left(\bJ_{\omega_i} \dtheta \right) ~^{0}\mathrm{\bI}_{\mathcal{A}_i}~ \bJ_{\omega_i} \right) \dtheta_r + \right. {\cdot} \cdot \cdot \\
&\hspace{2cm}\left. -m_i~S(^{0}\mathrm{\brho}_i)~\bg_0\right).
\end{split}
\end{align}
\end{subequations}

Analogous to Eq.~\eqref{eq:jacjacjac}, it is clear that
$^{0}\mathrm{\bomega}_{r_i}=\bJ_{\omega_i} \dtheta_r$ and
$^{0}\mathrm{\dbomega}_{r_i}=\bJ_{\omega_i} \ddtheta_r+
\bdJ_{\omega_i} \dtheta_r$. Furthermore, for SPRs,
it may be considered that $\dtheta_r={^{0}\mathrm{\bomega}_{r}}$ and $\ddtheta_r={^{0}\mathrm{\dbomega}_{r}}$. Hence, the following relation is derived by substituting
these formulas into Eq.~\eqref{eq:Slotine_Implicit}:
\begin{subequations}\label{eq:Slotine_Equality}
\begin{align}
&\bn_{S_i}= {\bJ_{\omega_i}}^{T} \left(^{0}\mathrm{\bI}_{\mathcal{A}_i}~^{0}\mathrm{\dbomega}_{r_i}+~S(^{0}\mathrm{\bomega}_i) \left(^{0}\mathrm{\bI}_{\mathcal{A}_i}~^{0}\mathrm{\bomega}_{r_i}\right)-m_i~S(^{0}\mathrm{\brho}_i)~\bg_0\right), \label{eq:Slotine_Equality1}\\
& \bn_{S_p}= \left(^{0}\mathrm{\bI}_{\mathcal{A}_p}\right) {^{0}\mathrm{\dbomega}_{r}} + \left(S\left({^{0}\mathrm{\bomega}}\right)~^{0}\mathrm{\bI}_{\mathcal{A}_p}\right) {^{0}\mathrm{\bomega}_{r}}-m_p~S\left (^{0}\mathrm{\brho}_p \right) \bg_0. \label{eq:Slotine_Equality2}
\end{align}
\end{subequations}

To derive the S-L regressor from equation~\eqref{eq:Slotine_Equality},
one may use
$^{0}\mathrm{\dbomega}_{r_i}=~^{0}\mathrm{\bR}_i~^{i}\mathrm{\dbomega}_{r_i}$,
$^{0}\mathrm{\bomega}_{r_i}=~^{0}\mathrm{\bR}_i~^{i}\mathrm{\bomega}_{r_i}$,
\textcolor{black}{$^{0}\mathrm{\bI}_{\mathcal{A}_i}=~^{0}\mathrm{\bR}_i~^{i}\mathrm{\bI}_{\mathcal{A}_i}~{^{0}\mathrm{\bR}_i}^T$,
and Eq.~\eqref{maping}}. Therefore, different parts of equation of
\eqref{eq:Slotine_Equality1} will be rewritten as follows:
\begin{subequations}\label{eq:Slotine_Implicit}
\begin{align}
&^{0}\mathrm{\bI}_{\mathcal{A}_i}~^{0}\mathrm{\dbomega}_{r_i}=~^{0}\mathrm{\bR}_i \left(~^{i}\mathrm{\bI}_{\mathcal{A}_i}~^{i}\mathrm{\dbomega}_{r_i}\right) \label{eq:OMG_Convert1}, \\
&S(^{0}\mathrm{\bomega}_i) \left(^{0}\mathrm{\bI}_{\mathcal{A}_i}~^{0}\mathrm{\bomega}_{r_i}\right)= ~^{0}\mathrm{\bR}_i \left(S(^{i}\mathrm{\bomega}_{i}) \left(^{i}\mathrm{\bI}_{\mathcal{A}_i}~^{i}\mathrm{ \bomega}_{r_i}\right)\right) \label{eq:OMG_Convert2}, \\
& -m_i~ S(^{0}\mathrm{\brho}_i)~\bg_0= -m_i~
^{0}\mathrm{\bR}_i~S\left(^{i}\mathrm{\brho}_i\right)~{^{0}\mathrm{\bR}_i}^T
~\bg_0 \label{eq:G_Convert1}.
\end{align}
\end{subequations}

Considering $^{i}\mathrm{\bg_0}=~{^{0}\mathrm{\bR}_i}^T~\bg_0$, and
using $S\left(^{i}\mathrm{\brho}_i\right)~^{i}\mathrm{\bg_0}=
-S\left(^{i}\mathrm{\bg_0}\right)~^{i}\mathrm{\brho}_i$,
\textcolor{black}{Eq.~\eqref{eq:G_Convert1} will be rewritten as:}
\begin{equation}\label{eq:G_Convert2}
-m_i~ S(^{0}\mathrm{\brho}_i)~\bg_0=~
^{0}\mathrm{\bR}_i~S\left(^{i}\mathrm{\bg_0}\right) \left(m_i~^{i}\mathrm{\brho}_i \right).
\end{equation}

A similar process may be applied to the derivation of the
\textcolor{black}{S-L} regression form of the moving platform
dynamic formulation. Notice that
$^{p}\mathrm{\dbomega}_r={{^{0}\mathrm{\bR}}_p}^T ~
^{0}\mathrm{\dbomega}_r$, $^{p}\mathrm{
\bomega}_r={^{0}\mathrm{\bR}_p}^T ~ ^{0}\mathrm{\bomega}_r$,
\textcolor{black}{$^{0}\mathrm{\bI}_{\mathcal{A}_p}=~^{0}\mathrm{\bR}_p~^{i}\mathrm{\bI}_{\mathcal{A}_p}~{^{0}\mathrm{\bR}_p}^T$},
and $^{p}\mathrm{\bg_0}={{^0}\mathrm{\bR}_p}^T~\bg_0$. Therefore,
the dynamic formulation of the SPRs with respect to the robot's
inertial parameters may be rewritten as:
\begin{subequations}\label{eq:implocal_S}
\begin{align}
&\bn_{S_i}={\bJ_{\omega_i}}^{T}~^{0}\mathrm{\bR}_i
\left(^{i}\mathrm{\bI}_{\mathcal{A}_i}~^{i}\mathrm{\dbomega}_{r_i}+
S(^{i}\mathrm{\bomega}_i)
\left(^{i}\mathrm{\bI}_{\mathcal{A}_i}~^{i}\mathrm{ \bomega}_{r_i}\right)+S\left(^{i}\mathrm{\bg_0}\right)\left(m_i~^{i}\mathrm{\brho}_i\right)\right), \label{eq:implocal_op_S}\\
& \bn_{S_p}=~^{0}\mathrm{\bR_{p}} \left(^{p}\mathrm{\bI}_{\mathcal{A}_p} ~^{p}\mathrm{\dbomega}_r\right)+S\left({^{0}\mathrm{\bomega}}\right) {^{0}\mathrm{\bR_{p}}} \left(^{p}\mathrm{\bI}_{\mathcal{A}_p} ~^{p}\mathrm{ \bomega}_r\right)+S\left(^{p}\mathrm{\bg_0}\right) \left(m_p~^{p}\mathrm{\brho}_p\right). \label{eq:implocal_oi_S}
\end{align}
\end{subequations}

On the other hand, in order to transform the first and second terms of
equations of \eqref{eq:implocal_op_S} and \eqref{eq:implocal_oi_S}
to linear forms, it may be proved that
\cite{danaei2017dynamic,codourey1997body}:
\begin{subequations}\label{eq:mapcod_S}
\begin{align}
& {^{i}\mathrm{\bI}_{\mathcal{A}_i}}~^{i}\mathrm{\dbomega}_{r_i}=~^{i}\mathrm{\dTbomega}_{r_i}~^{i}\mathrm{\BI}_{\mathcal{A}_i},\\
&{^{i}\mathrm{\bI}_{\mathcal{A}_i}}~^{i}\mathrm{\bomega}_{r_i}= ~^{i}\mathrm{\Tbomega}_{r_i}~^{i}\mathrm{\BI}_{\mathcal{A}_i},
\end{align}
\end{subequations}
in which, $^{i}\mathrm{\BI}_{\mathcal{A}_i}$ and $^{i}\mathrm{\Tbomega}_i$ are
defined as:
\begin{subequations}
\begin{align}
& ^{i}\mathrm{\BI}_{\mathcal{A}_i}=[I_{xx\mathcal{A}_i},I_{xy\mathcal{A}_i},I_{xz\mathcal{A}_i},I_{yy\mathcal{A}_i},I_{yz\mathcal{A}_i},I_{zz\mathcal{A}_i}],\\
&^{i}\mathrm{\Tbomega}_{r_i}=\begin{bmatrix}
{\omega_r}_{x_i} & {\omega_r}_{y_i} & {\omega_r}_{z_i} & 0 & 0 & 0 \\
0 & {\omega_r}_{x_i} & 0 & {\omega_r}_{y_i} & {\omega_r}_{z_i} & 0\\
0 & 0 & {\omega_r}_{x_i} & 0 & {\omega_r}_{y_i} & {\omega_r}_{z_i}
\end{bmatrix},
\end{align}
\end{subequations}

Now, substitute \eqref{eq:mapcod_S} into \eqref{eq:implocal_S}. The
linear S-L regressor form of SPRs with respect to
the inertial parameters may be derived as:
\begin{equation}\label{regressorform_Slotine}
{\bJ_\omega}^{T} \btau=\textcolor{black}{\bn_S+\bn_d}=\bY_{S}(\btheta,\dtheta,\dtheta_r,\ddtheta_r) \bpi,
\end{equation}
in which, \textcolor{black}{the S-L regressor $\bY_S$} is defined as
follows:
\begin{subequations}\label{eq:Slotine_Regressor}
\begin{align}
&\bY_{S}=\begin{bmatrix}
\bY_{S_{p}} & ,\bY_{{S}_i} & ,... &,\bY_{S_{n}}
\end{bmatrix},\\
& \bY_{S_p}= \begin{bmatrix}
^{0}\mathrm{\bR}_p~S\left(^{p}\mathrm{\bg_0}\right) &,\left(^{0}\mathrm{\bR}_p~
^{p}\mathrm{\dTbomega}_{r}+S\left({^{0}\mathrm{\bomega}}\right) ~^{0}\mathrm{\bR}_p~^{p}\mathrm{\Tbomega}_{r}\right)
\end{bmatrix}, \label{eq:Slotine_Regressor_MP}\\
& \bY_{S_i}= {\bJ_{\omega_i}}^{T}~^{0}\mathrm{\bR}_i
\begin{bmatrix}
S\left(^{i}\mathrm{\bg_0}\right) & ,
\left(^{i}\mathrm{\dTbomega}_{r_i}+ S(^{i}\mathrm{\bomega}_i)~^{i}\mathrm{\Tbomega}_{r_i}\right)
\end{bmatrix}.
\end{align}
\end{subequations}
and the inertial parameters $\bpi$ are defined as:
\begin{subequations}\label{eq:pi}
\begin{align}
& \bpi= \begin{bmatrix}
\bpi_p & \bpi_i & ... & \bpi_n
\end{bmatrix}^T~,~\in \R^{9k},\\
& \bpi_p= \begin{bmatrix}
m_p~^{p}\mathrm{\brho}_p & ^{p}\mathrm{\BI}_{\mathcal{A}_p}
\end{bmatrix}^T, \\
& \bpi_i= \begin{bmatrix}
m_i~^{i}\mathrm{\brho}_i & ^{i}\mathrm{\BI}_{\mathcal{A}_i}
\end{bmatrix}^T.
\end{align}
\end{subequations}

A careful examination of Eq.~\eqref{eq:pi}, confirms that, in SPRs,
the first moment and the moment of inertia are the only inertial
parameters in vector $\bpi$, so that the pure masses of the links do
not directly contribute to the dynamics. \textcolor{black}{Moreover,
in order to derive linear regressor form, it is enough to substitute
$\ddtheta_r=\ddtheta$ and $\dtheta_r=\dtheta$ in
Eq.~\eqref{eq:Slotine_Regressor}, which leads to
$^{p}\mathrm{\dTbomega}_{r}=~^{p}\mathrm{\dTbomega}$ and
$^{p}\mathrm{\Tbomega}_{r}=~^{p}\mathrm{\Tbomega}$, as well as
$^{i}\mathrm{\dTbomega}_{r_i}=~^{i}\mathrm{\dTbomega}_{i}$ and
$^{i}\mathrm{\Tbomega}_{r_i}=~^{i}\mathrm{\Tbomega}_{i}$}.

\subsection{Unique Regressor Reduction} \label{Reduction}
\textcolor{black}{Careful examinations of regressor forms stated in
Eq.~\eqref{eq:Slotine_Regressor} reveal the stringent requirement of
order reduction to eliminate the zero or linearly dependent elements
in the regressor. In practice, although using a reduced regressor in
adaptive model-based controllers does not improve the performance of the controller, it significantly decreases the required real-time computational process. On the other hand, in the calibration schemes
in which linear regression form of robot dynamics is used, it is
inevitable to use a reduced regressor to avoid having a
rank-deficient observation matrix and to make elements of the
inertial parameters fully identifiable. }

\textcolor{black}{For this means, we want to eliminate the zero or
linearly dependent elements in the obtained regressor.}
\textcolor{black}{For this purpose, the method mentioned in
Ref.~\cite{klodmann2015closed}, which has a unique solution, is
used.} To this end, one may represent the observation matrix of SPRs
as:
\begin{equation}
\bW(\btheta,\dtheta,\ddtheta) =\begin{bmatrix}
\bY(t_1)^T & ... & \bY(t_k)^T
\end{bmatrix}
^T ~,~\in \R^{jm\times 9k},
\label{eq:Delta}
\end{equation}
in which $j$ represents the number of observations.
\textcolor{black}{If $j$ is sufficiently large, the condition $jm
\geq 9k$ shall be met for the system to be observable}. Hence, the
reduced regressor matrix $\bY_r$ and the \textcolor{black}{BIPs}
$\bpi_r$ shall be found such that to preserve the full rank $p$ of
the observation matrix $W$, in order for the system to be
observable. In such a case, it may be proved that there is a unique
transformation matrix $\bB \in \R^{p \times 9 k}$ and a matrix
$\bS$, spanning the row space of the observation matrix such that
there exist a generalized inverse $\psB=\bS \bB^{T}\left(\bB\bS
\bB^{T}\right)^{-1}$.
\textcolor{black}{In order to derive the $\bB$ matrix, it suffices
to calculate the reduced-row echelon form of $\bW$, which is the
upper triangular matrix $\bB_E$. By eliminating the zero rows of
$\bB_E$, the matrix $\bB$ is derived.} By this means, $\bY_{r} \in
\R^{m\times p} $ and $\bpi_r \in \R^{p}$ are derived as:
\begin{subequations}\label{eq:Reduced}
\begin{align}
&\bY_r=\bY \psB, \\
&\bpi_r=\bB \bpi.
\end{align}
\end{subequations}
{It should be noted that, the $\bB_E$ matrix is
only a function of the kinematic parameters of the robot, and has
numerical values. Thus, to calculate it, $rref$ command in MATLAB
can be used for the above mentioned calculations.}

\section{Examples}\label{sec:Examples}

\textcolor{black}{To verify the validity of the formulation
presented in this paper on different SRRs, the literature was
re-examined to select suitable case studies. According to different
kinematic architectures, some of the notable SPRs may be listed as:}

\begin{itemize}
\item \textcolor{black}{\textbf{2-DOF SPRs:}
SPRs with 5R structures, their various configurations have been
widely mentioned in the literature
\cite{Safaryazdi_2017,danaei2017dynamic}. In addition, SPRs with as
7-bars \cite{Nie_2020}, with 8-bars \cite{Wiitala_1998},
RR-\underline{R}\underline{R}R-RRR \cite{Kong_2010}, and
2-\underline{R}RR \cite{Duan_2016} may be listed.}

\item \textcolor{black}{\textbf{3-DOF SPRs:}
SPRs with 3-\underline{R}RR structures
\cite{gosselin1994agile,Li_2019},
3-\underline{R}RP \cite{Akbarzadeh_2010},
3-\underline{R}RS \cite{Du_2017},
and 3-P\underline{R}R \cite{Zarkandi_2021}.}
\end{itemize}

\textcolor{black}{In this paper, two distinguished SPRs, namely, the
ARAS-Diamond (2-DOF 5R) \cite{Ebr}, and the standard
3-\underline{R}RR, are selected as the case studies in this paper.
The reasons for choosing these robots are as following:}

\begin{enumerate}
\item \textcolor{black}{Evaluation of SPRs with different kinematic structures,
one example with a 5R structure and the other with a
3-\underline{R}RR structure. Kinematic analysis of these robots will
show how Jacobian matrices and their derivatives (in fact, the terms
of angular velocity and acceleration) for each link can be derived.}

\item \textcolor{black}{Investigation of different approaches to the
description of the orientation of the moving platform in the task space.
It will be shown that the use of Euler angles requires minor modifications
to the proposed dynamic formulation.}
\end{enumerate}

\textcolor{black}{In the next section, the kinematic analysis of the
two case studies is examined.}
\subsection{ARAS-Diamond Robot}\label{Ex:Diamond}
The ARAS-Diamond robot is a \textcolor{black}{PR} developed to
perform the minimally invasive vitreoretinal eye surgery. According
to the robot's spherical structure, all the links have a pure
rotational motion about the remote center of motion (RCM).
\textcolor{black}{The kinematic details of this robot are given in
details in \cite{Ebr}, and in this article, only its important
issues are reviewed.}
\subsubsection{Robot Geometry}
\begin{figure}[!ht]
\includegraphics[scale=0.7]{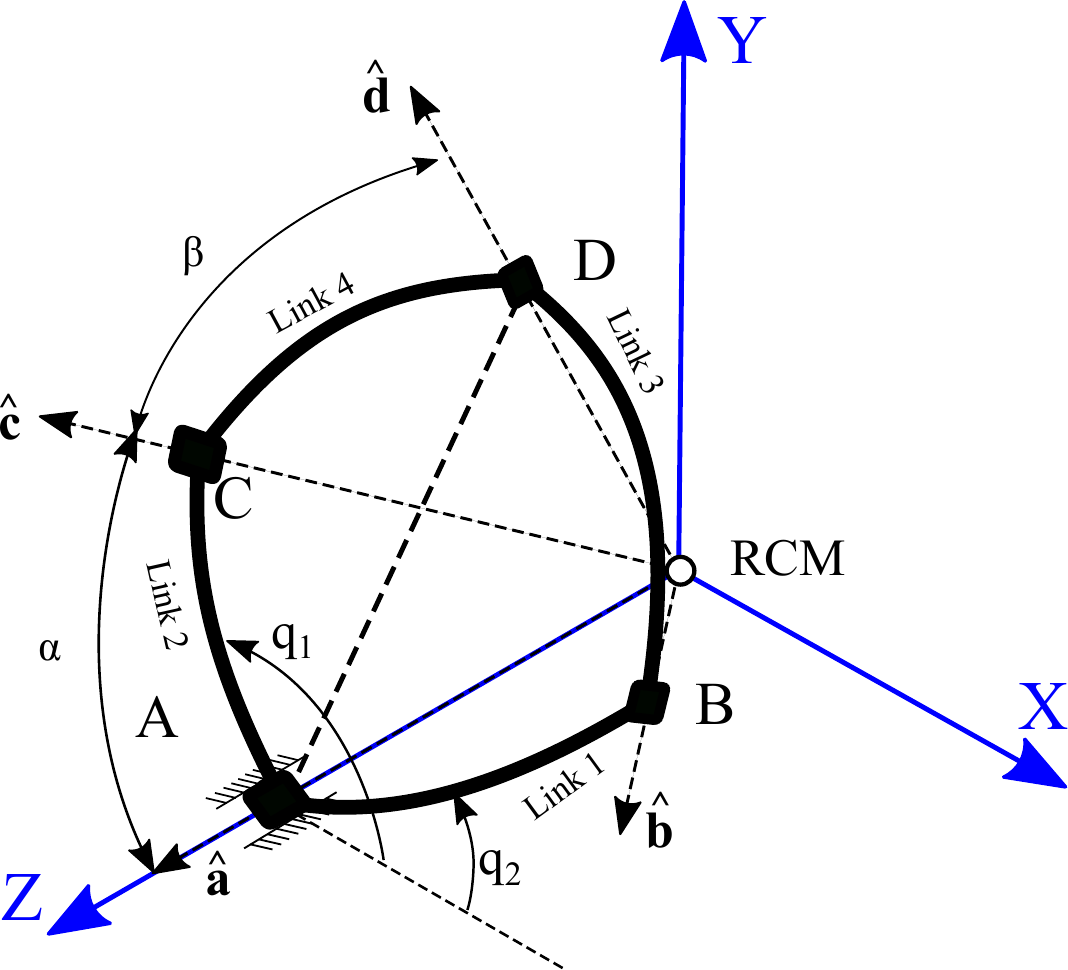}
\caption{ Schematic of the ARAS-Diamond SPR} \label{fig:Sch}
\end{figure}

The kinematic structure of the ARAS-Diamond SPR is depicted in
Fig.~\ref{fig:Sch}. 
Similar to
\cite{Ebr}, the position of this point in the spherical coordinate system is considered as the generalized coordinate.
\textcolor{black}{Task space variables $\btheta=[\phi,\gamma]^T$} are sufficient to describe the position of this point, since the ARAS-Diamond robot structure is designed such that
all the links are enclosed in a sphere. As a result, adopting
spherical coordinates instead of Cartesian coordinates leads to a
simpler kinematic analysis~\cite{Ebr}.

As illustrated in Fig.~\ref{fig:Sch}, the geometric parameters of the
robot may be specified by $\alpha$ and $\beta$, and the four unit-vectors
of $\hba$, $\hbb$, $\hbc$, and $\hbd$ represent the directions from the RCM
point to the revolute joints of the robot \textcolor{black}{as}:
\begin{subequations}
\begin{align}
&\hba=[0,0,1]^T ~~, ~~ \hbb=\bR_{z} \left(q_2+\frac{\pi}{2}\right) \bR_{x}(\alpha)~\hba, \\
&\hbc=\bR_{z} \left(q_1+\frac{\pi}{2}\right) \bR_{x}(\alpha)~\hba ~~, ~~ \hbd=\bR_{z}(\phi) \bR_{y}(\gamma) \hba.
\end{align}
\end{subequations}

\subsubsection{Differential Kinematic Analysis}\label{jacoo}
\textcolor{black}{The inverse kinematics of the ARAS-Diamond robot may be expressed as follows \cite{Ebr}:}
\begin{subequations}\label{invK_2}
\begin{align}
q_1 &= \phi+ \arccos{\left(\frac{\cos{\beta}-\cos{\gamma}\cdot\cos{\alpha}}{\sin{\gamma}\cdot\sin{\alpha}}\right)},\\
q_2 &= \phi -\arccos{\left(\frac{\cos{\beta}-\cos{\gamma}\cdot\cos{\alpha}}{\sin{\gamma}\cdot\sin{\alpha}}\right)}.
\end{align}
\end{subequations}
In order to derive the Jacobian matrix, the time derivative of
\eqref{invK_2} is computed. Accordingly, $\dot q_1$ and
$\dot q_2$ may be represented as:
\begin{subequations}\label{eq:VelJac}
\begin{align}
&\dot{q}_1 = \dot \phi + \textcolor{black}{h}~ \dot \gamma \label{eq:thetadot1},\\
&\dot{q}_2 = \dot \phi - \textcolor{black}{h}~ \dot \gamma, \label{eq:thetadot2}
\end{align}
\end{subequations}
where, \textcolor{black}{scalar parameter $h$} is given in Appendix \ref{Appendix: Diamond}. Consequently, by arranging \eqref{eq:VelJac} into a matrix form, the Jacobian matrix is given as:
\begin{equation}\label{eq:Jacobian}
\bJ_\omega=\begin{bmatrix}
1 & +h\\
1 & -h
\end{bmatrix}.
\end{equation}

In order to derive passive joint Jacobians, which map the velocity from the task-space to the passive joints $\dot \bq_{p}=[\dot
{\hat{C}},\dot {\hat{B}}]^T$, we must compute $\dot {\hat{C}}$
and $\dot {\hat{B}}$. \textcolor{black}{It can be proved that:}
\begin{equation}\label{eq:Bdot}
\dot {\hat{B}}=- \dot {\hat{C}} = \textcolor{black}{s}~\dot \gamma,
\end{equation}
in which, \textcolor{black}{the scalar parameter $s$ } is given in Appendix \ref{Appendix: Diamond}.
As a result, the Jacobian matrix of the passive angles, is obtained as:
\begin{equation}\label{Jacop}
\bJ_{\omega_p}=\begin{bmatrix}
0 & -\textcolor{black}{s} \\ 0 & +\textcolor{black}{s}
\end{bmatrix}.
\end{equation}

\subsubsection{Jacobian Analysis of Each Link}

In order to calculate the angular velocity of each link of the robot, it is split into two identical serial arms.
Hence, the angular velocity of each link may be derived as follows:
\begin{subequations}\label{eq:omega}
\begin{align}
&^{0}\mathrm{\bomega_1}= \dot q_2 \cdot \hba, \label{eq:Omega1}\\
&^{0}\mathrm{\bomega_2}= \dot q_1 \cdot \hba, \label{eq:Omega2}\\
&^{0}\mathrm{\bomega_3}= \dot q_2 \cdot \hba - \dot {\hat{B}} \cdot \hbb=~^{0}\mathrm{\bomega_1}- \dot {\hat{B}} \cdot \hbb, \label{eq:Omega3} \\
&^{0}\mathrm{\bomega_4}= \dot q_1 \cdot \hba + \dot {\hat{C}} \cdot \hbc= ~^{0}\mathrm{\bomega_2}+ \dot {\hat{C}} \cdot \hbc. \label{eq:Omega4}
\end{align}
\end{subequations}
Therefore, the Jacobian matrix of each link is given \textcolor{black}{by}:
\begin{equation}\label{eq:Jacobian1}
\bJ_{\omega_1}=\begin{bmatrix} \hba & -h\hba \end{bmatrix}, ~~ \bJ_{\omega_2}=\begin{bmatrix}
\hba & +h\hba
\end{bmatrix},
\end{equation}
\begin{equation}\label{eq:Jacobian2}
\bJ_{\omega_3}=\bJ_{\omega_1}+\begin{bmatrix} \textbf{0}_{3 \times 1} & -s\hbb \end{bmatrix}, ~~ \bJ_{\omega_4}=\bJ_{\omega_2}+\begin{bmatrix} \textbf{0}_{3 \times 1} & -s\hbc \end{bmatrix},
\end{equation}
in which, $\textbf{0}_{3 \times 1}$ denotes the $3 \times 1$ zero vector.
\subsubsection{Acceleration Analysis}

The derivative of the Jacobian matrix of the first and second links are easily be obtained as:
\begin{equation}\label{eq:Jacobiandot1}
\bdJ_{\omega_1}=\begin{bmatrix} \textbf{0}_{3 \times 1} & -\dot h \hba \end{bmatrix},~~\bdJ_{\omega_2}=\begin{bmatrix} \textbf{0}_{3 \times 1}& +\dot h\hba \end{bmatrix},
\end{equation}
\textcolor{black}{where $\dot h= c~\dot \gamma$, and the scalar parameter $c$ is given in Appendix \ref{Appendix: Diamond}.}
Furthermore, the derivative of the Jacobian matrices of the
third and fourth links, requires the time derivatives of unit vectors
$\hbc$ and $\hbd$ and the scalar parameter $s$. The time derivative of $s$ is computed using Maple
symbolic manipulation, which yields the expression of Eq.~\eqref{eq:Ali}. Thence, the derivative of the Jacobian matrices of the third and fourth links are obtained \textcolor{black}{as}:
\begin{equation}\label{eq:jacobian3dot}
\bdJ_{\omega_3}=\bdJ_{\omega_1} + \begin{bmatrix} \textbf{0}_{3 \times 1} & -\left(\dot s \cdot\hbb+ s \cdot \dhbb \right) \end{bmatrix}, ~~ \bdJ_{\omega_4}=\bdJ_{\omega_2} + \begin{bmatrix}
\textbf{0}_{3 \times 1} & -\left(\dot s \cdot \hbc+ s \cdot \dhbc\right)
\end{bmatrix}.
\end{equation}


\subsection{The 3-\underline{R}RR Spherical Parallel Manipulator}\label{Ex:3RRR_Robot}
\begin{figure}[!ht]
\includegraphics[scale=.7]{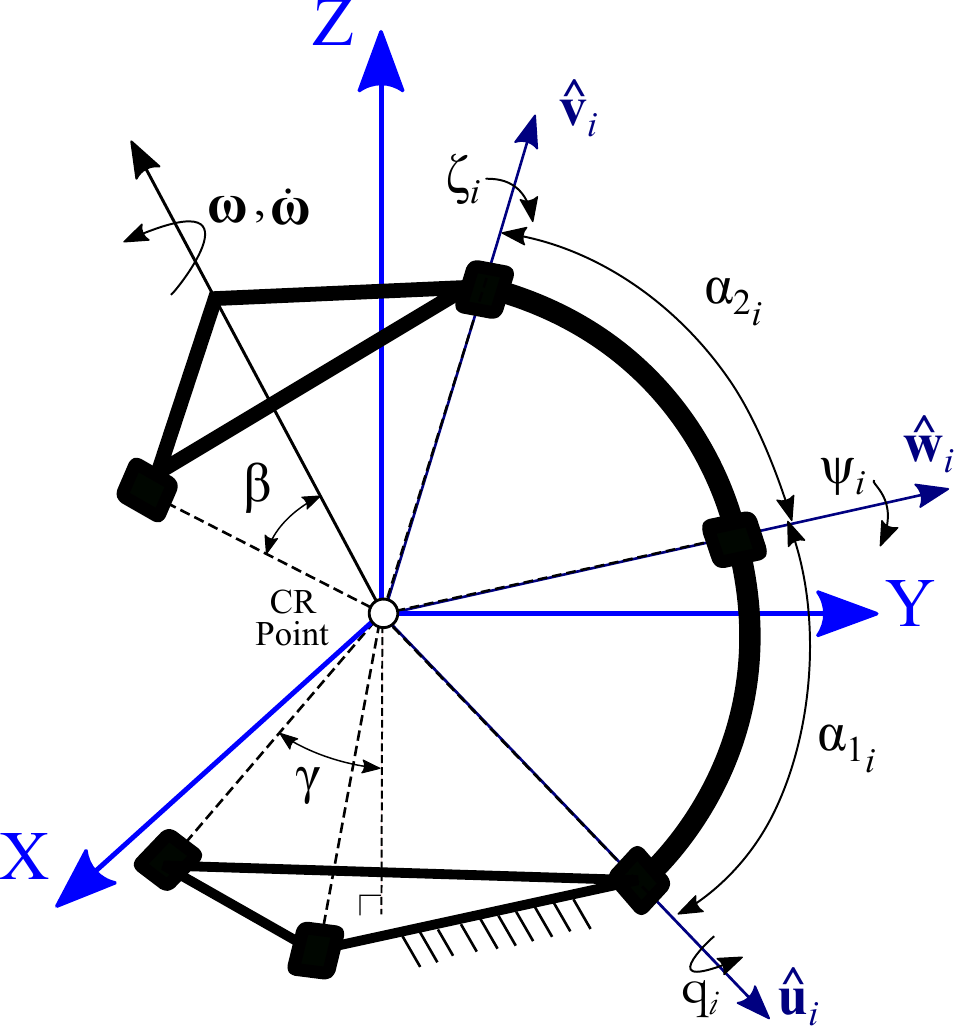}
\caption{ Schematic of a typical 3-\underline{R}RR spherical parallel manipulator} \label{fig:General_Sch2}
\end{figure}

The 3-\underline{R}RR spherical parallel manipulator is a
symmetrical 3-DOF mechanism composed of three identical kinematic
chains. Each kinematic chain is itself composed of a proximal and a
distal link, and three revolute joints. The proximal links are
driven by actuators, while distal links and the moving platform are
connected together with passive revolute joints. The structure of
the 3-\underline{R}RR manipulator is depicted in
Fig.~\ref{fig:General_Sch2}. \textcolor{black}{The kinematic
characteristics of this robot are mentioned in details in
\cite{abedloo2014closed}, and in this article, only its important
issues are reviewed.}

\subsubsection{Robot Geometry}


For kinematic analysis of the 3-\underline{R}RR spherical parallel manipulator,
the orientation of the moving platform is described using $ZYX$ fixed-body
Euler angles. Therefore, $\btheta=[\theta_1,\theta_2,\theta_3]^T$ are
considered as the task space variables,
\textcolor{black}{Thus, the rotation matrix of the moving
platform may be derived as $^{0}\mathrm{\bR}_p=\bR_z(\theta_1)\bR_y(\theta_2)\bR_x(\theta_3)$.}

In the 3-\underline{R}RR manipulator, unit vectors $\hbu_i$,
$\hbv_i$, and $\hbw_i$, for $\rm i=1,2,3$, all intersect at the CR point, which is defined as the origin of the base
coordinate system. Therefore, the unit vector of the primary and the
intermediate revolute joints can be formulated as:
\begin{subequations}\label{eq:Unit_Vectors_UW}
\begin{align}
& \hbu_i=\bR_z(\lambda_i)\bR_x(\gamma-\pi) [0,0,1]^T,\\
&\hbw_i=\bR_z(\lambda_i)\bR_x(\gamma-\pi)\bR_z(q_i)\bR_x(\alpha_{1_i})[0,0,1]^T. \label{eq:Unit_Vector_W}
\end{align}
\end{subequations}
Moreover, $\hbv_i$ may be derived as follows:
\begin{subequations}\label{eq:Unit_Vectors_V}
\begin{align}
&{{\hbv}_i}^*=\bR_z(\eta_i)\bR_x(-\beta) [0,0,1]^T,\\
&\hbv_i=~^{0}\mathrm{\bR}_p~{{\hbv}_i}^*,
\end{align}
\end{subequations}
in which, ${\hbv_i}^*$ represents the unit vector ${\hbv_i}$ in the initial
configuration of the moving platform.

\subsubsection{Differential Kinematic Analysis}
Before beginning the Jacobian analysis of the 3-\underline{R}RR
manipulator, it should be noted that by choosing the Euler angles as a way
to describe the moving platform orientation, the angular velocity of the moving platform would not be equal to the rate of change of the Euler
angles, $\dtheta=[\dot \theta_1,\dot \theta_2,\dot \theta_3]^T$. Instead,
this relationship between these two vectors would be defined as:
\begin{equation}\label{eq:Euler_Mapping_3RRR}
{^{0}\mathrm{\bomega}}=\bE \dtheta,
\end{equation}
in which, $\bE(\theta_1,\theta_2,\theta_3)$ relates the
rates of the Euler angles to the angular velocity of the moving platform. In the $ZYX$ fixed body Euler angles, this matrix is given by:
\begin{equation}
\bE=\begin{bmatrix}
0 & -\sin(\theta_1) & \cos(\theta_1) \cos(\theta_2) \\
0 & \cos(\theta_1) & \sin(\theta_1) \cos(\theta_2)\\
1 & 0 & -\sin(\theta_2)
\end{bmatrix}.\\
\end{equation}
Therefore, the angular velocity of the moving platform may be
derived as the sum of its successive angular rotations as:
\begin{equation}\label{eq:SumVel_3RRR}
{^{0}\mathrm{\bomega}}= \dot q_i \cdot \hbu_i+\dot \psi_i \cdot \hbw_i + \dot \zeta_i \cdot \hbv_i.
\end{equation}
\textcolor{black}{In this equation, $\dot q_i$ and $\dot \psi_i$ are derived by dot-multiplying both
sides of the equation of \eqref{eq:SumVel_3RRR} in $\left(\hbv_i \times
\hbw_i \right)$ and $\left( \hbu_i \times \hbv_i \right)$, respectively, which
results in:}
\begin{subequations}\label{eq:dot_3RRR}
\begin{align}
&\dot q_i=\frac{\left(\hbv_i \times \hbw_i\right)}{\left(\hbv_i \times \hbw_i\right) \cdot \hbu_i}\cdot {^{0}\mathrm{\bomega}} ~\textcolor{black}{=\bJ_{\omega_i} ~\dtheta}~~,~~ \rm for~~ i=1,2,3, \label{eq:dot_q_3RRR}\\
&\dot \psi_i=\frac{\left(\hbu_i \times \hbv_i\right)}{\left(\hbu_i \times \hbv_i\right) \cdot \hbw_i}\cdot {^{0}\mathrm{\bomega}} ~\textcolor{black}{= \bJ_{\omega_{p_i}} ~\dtheta}~~,~~ \rm for~~i=1,2,3. \label{eq:dot_psi_3RRR}
\end{align}
\end{subequations}
Now, in order to derive the Jacobian \textcolor{black}{matrices} from equations of
\eqref{eq:dot_q_3RRR} and \eqref{eq:dot_psi_3RRR}, the following
parameters are defined for the sake of simplification:
\begin{subequations}\label{eq:Extra_Parameters_3RRR}
\begin{align}
&\bp_{1_i}=\hbv_i \times \hbw_i,\\
&\bp_{2_i}=\hbu_i \times \hbv_i,\\
&h_{i}=\left(\hbv_i \times \hbw_i \right)\cdot \hbu_i= \bp_{1_i} \cdot \hbu_i.
\end{align}
\end{subequations}
Finally, using equations of \eqref{eq:Euler_Mapping_3RRR},
\eqref{eq:dot_3RRR}, and \eqref{eq:Extra_Parameters_3RRR}, each row
of the input-output Jacobian and the passive Jacobian matrices are derived as:
\begin{subequations}
\begin{align}
&\bJ_{\omega_i}~={h_i}^{-1}~{{\bp}_{1_i}}^T~\bE,~~ \rm{for}~~i=1,2,3, \label{eq:Ji-3RRR}\\
& \bJ_{{\omega_p}_i} = {h_i}^{-1}~{{\bp}_{2_i}}^T~\bE,~~ \rm{for}~~i=1,2,3. \label{eq:Jp-3RRR}
\end{align}
\end{subequations}

\subsubsection{Jacobian Analysis of Links}
The angular velocity of the \textcolor{black}{proximal and distal links} of each
kinematic chain may be written as:
\begin{subequations}
\begin{align}
&^{0}\mathrm{\bomega}_{1_i}
= \dot q_i \cdot \hbu_i ~\textcolor{black}{= \bJ_{\omega_{1_i}} ~ \dtheta}, \\
&^{0}\mathrm{\bomega}_{2_i}= ~^{0}\mathrm{\bomega}_{1_i} + \dot \psi_i \cdot \hbw_i ~\textcolor{black}{= \bJ_{\omega_{2_i}} ~ \dtheta}.
\end{align}
\end{subequations}
Therefore, using equations \eqref{eq:dot_q_3RRR} and
\eqref{eq:dot_psi_3RRR}, each row of the Jacobian matrix of the
proximal and distal links are formulated as follows:
\begin{subequations}
\begin{align}
&\bJ_{\omega_{1_i}}=
{\hbu_i}~\bJ_{\omega_i}, \label{eq:Jw1_3RRR}
\\
&\bJ_{\omega_{2_i}}= \bJ_{1_i}+ {\hbw_i}~\bJ_{\omega_{p_i}}. \label{eq:Jw2_3RRR}
\end{align}
\end{subequations}

\subsubsection{Acceleration Analysis}
In order to derive the acceleration terms in the proposed dynamic
formulation, it is necessary to obtain the time derivative of $\dot q_i$
and $\dot \psi_i$. Thus, the time derivative of each row of the
corresponding Jacobian matrices $\bJ_\omega$ and $\bJ_{\omega_p}$ is computed symbolically as:
\begin{equation}
\bdJ_{\omega_i}= \frac{{\dbp_{1_i}}^T \bE+{\bp_{1_i}}^T \dbomega}{h_i} - \frac{{\bp_{1_i}}^T \bE~\dot{h}_{i}}{{h_{i}}^2},
\end{equation}
\begin{equation}
\bdJ_{{\omega_p}_i}= \frac{ {\dbp_{2_i}}^T \bE+{\bp_{2_i}}^T \dbE}{h_i} - \frac{{\bp_{2_i}}^T \bE~\dot{h}_{i}}{{h_{i}}^2},
\end{equation}

In these equations, the method of calculating the time derivatives of vectors of
$\bp_{1_i}$ and $\bp_{2_i}$, and the value of scalar ${h}_{i}$ is given in Appendix \ref{Appendix:3RRR_Kinematic}. Therefore, the time
derivative of the Jacobian matrix of the proximal and the distal links may
be represented by:
\begin{subequations}
\begin{align}
&\bdJ_{\omega_{1_i}}= \hbu_i~\bdJ_{\omega_i}, \label{eq:dJW1_3RRR}
\\
&\bdJ_{\omega_{2_i}}= \bdJ_{\omega_{1_i}}+ \left(\dhbw_i ~\bJ_{\omega_{p_i}}+ \hbw_i~\bdJ_{\omega_{p_i}}\right), \label{eq:dJw2_3RRR}
\end{align}
\end{subequations}
in which, $\dhbw_i$ is given in the Appendix \ref{Appendix:3RRR_Kinematic}.

\subsubsection{Dynamic Formulations}
As mentioned earlier, the angular velocity of the moving platform is
related to the Euler angles rate with ${^{0}\mathrm{\bomega}}=\bE \dtheta$. In the
Jacobian analysis of each link, the effect of the $\bE$ matrix has been
considered. Thus, the terms related to the dynamics of the moving platform
given in Eq.~\eqref{eq:Implicit_Dy_Op}, are rewritten as follows:
\begin{equation}\label{eq:Euler_Dynamic}
{\mbJ_{\omega}}^T \btau= \left(^{0}\mathrm{\bI}_{\mathcal{A}_p}\right) {^{0}\mathrm{\dbomega}}+S\left({^{0}\mathrm{\bomega}}\right)\left(^{0}\mathrm{\bI}_{\mathcal{A}_p}\right) {^{0}\mathrm{\bomega}} -m_i~S\left (^{0}\mathrm{\brho}_p \right) \bg_0,
\end{equation}
in which each row of the Jacobian matrix ${\mbJ_{\omega}}$ is defined by ${\mbJ_{\omega}}_i={h_i}^{-1}~{{\bp}_{1_i}}^T$. Although the Eq.~\eqref{eq:Euler_Dynamic} has an explicit form, it is written
based on ${^{0}\mathrm{\bomega}}$ and ${^{0}\mathrm{\dbomega}}$ terms, while explicit dynamics
should be in terms of $\dtheta$ and $\ddtheta$ to have the same
force distribution as the dynamics of each link. As a result, using
the properties of $^{0}\mathrm{\bomega}=\bE \dtheta$ and ${^{0}\mathrm{\dbomega}}=\bE \ddtheta+ \dbE \dtheta$, and multiplying $\bE^{T}$ to the left
side of the Eq.~\eqref{eq:Euler_Dynamic}, the explicit dynamic of
the moving platform based on $\dtheta$ and $\ddtheta$, may be
written as follows:
\begin{equation}\label{eq:Euler_Dynamic_2}
{\bJ_\omega}^{T}\btau = \bM_p(\btheta) \ddtheta+\bC_p(\btheta,\btheta) \dtheta+\bg_p(\btheta).
\end{equation}
In this representation, each row of the Jacobian matrix
$\bJ_{\omega_i}=\mbJ_{{\omega}_i} \bE,~\rm{for}~i=1,2,3$, is equal
to each row of the Jacobian matrix of the whole manipulator, given
in equation \eqref{eq:Ji-3RRR}. Moreover, the explicit dynamic of
the moving platform may be represented by the following matrices:
\begin{subequations}\label{eq:Explicit_Euler}
\begin{align}
&\bM_p=\bE^T\left(^{0}\mathrm{\bI}_{\mathcal{A}_p}\right)\bE,\\
&\bC_p=\bE^T\left(^{0}\mathrm{\bI}_{\mathcal{A}_p}\right) \dbE+\bE^T \left(S(\bE \dtheta)\left(^{0}\mathrm{\bI}_{\mathcal{A}_p}\right)\right) \bE,\\
&\bg_p=-\bE^T\left( m_i~S\left (^{0}\mathrm{\brho}_p \right) \bg_0 \right).
\end{align}
\end{subequations}

The same process must be followed for the derivation of the
corresponding terms related to the moving platform in the
\textcolor{black}{regressors}. \textcolor{black}{Therefore, it is
enough to use $^{0}\mathrm{\bomega}_r=\bE \dtheta_r$ and
$^{0}\mathrm{\dbomega}_r=\bE \ddtheta_r+ \dbE \dtheta_r$ and
multiply the left side of equation \eqref{eq:Slotine_Regressor_MP}
by $\bE^T$}.


\section{Verification}

In order to verify
the dynamic formulation proposed in this paper, two sides of the following
equations are compared to each other for a typical trajectory $\btheta$:
\begin{subequations}\label{eq:Verification_Total}
\begin{align}
&\bM(\btheta)\ddtheta~+\bC(\btheta,\dtheta) \dtheta~+\bg(\btheta)=\bY(\btheta,\dtheta, \ddtheta) \bpi=\bY_r(\btheta,\dtheta, \ddtheta) \bpi_r, \label{eq:Ver_Reg} \\
&\bM(\btheta)\ddtheta_r+ \bC(\btheta,\dtheta) \dtheta_r+\bg(\btheta)=\bY_S(\btheta,\dtheta,\dtheta_r, \ddtheta_r) \bpi=\bY_{S_r}(\btheta,\dtheta,\dtheta_r, \ddtheta_r) \bpi_r. \label{eq:Ver_Reg_Slotine}
\end{align}
\end{subequations}

\begin{figure}[!htb]
\centering
\subfloat[ARAS-Diamond \label{fig:ADAMS1}]{\includegraphics[width=2.225in]{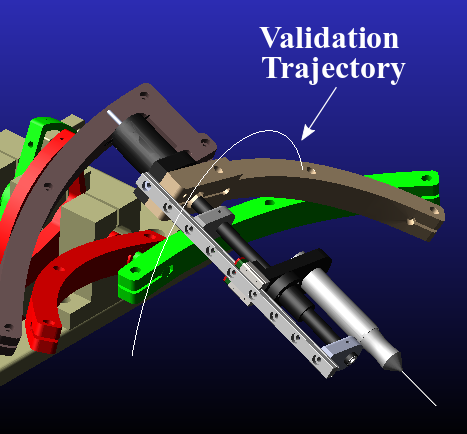}}
\hfil
\subfloat[3-\underline{R}RR \label{fig:ADAMS2}]{\includegraphics[width=2.1in]{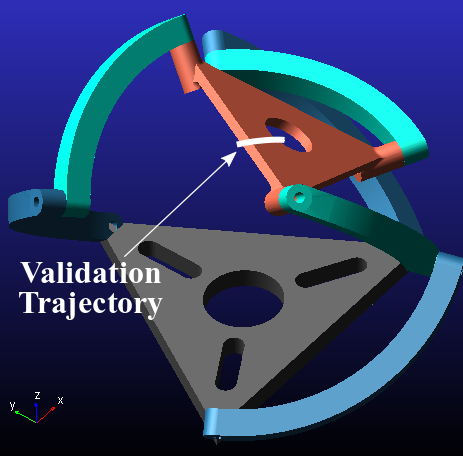}}
\caption{Considered Models in MSC-ADAMS\textsuperscript{\textregistered} }
\label{fig:ADAMS}
\end{figure}

\textcolor{black}{For the verification process, each robot's
explicit dynamics is both coded in MATLAB and modeled in the
MSC-ADAMS\textsuperscript{\textregistered} package. In the latter
case, a CAD model of the corresponding robot is used, as shown in
Fig.~\ref{fig:ADAMS1} and Fig.~\ref{fig:ADAMS2}. By applying the
desired trajectory $\btheta=\btheta_d$ to both models separately and
comparing the resulting output torques of the actuators, one may
conclude that the dynamic formulation is verified if the differences
are negligible.}
%


\subsection{ARAS-Diamond Robot Verification}\label{Ver:ARAS-Diamond}
\begin{figure}[!b]
\centering
\subfloat[Actuator torques computed through two models \label{fig:Dynamic}]{\includegraphics[width=2.3in]{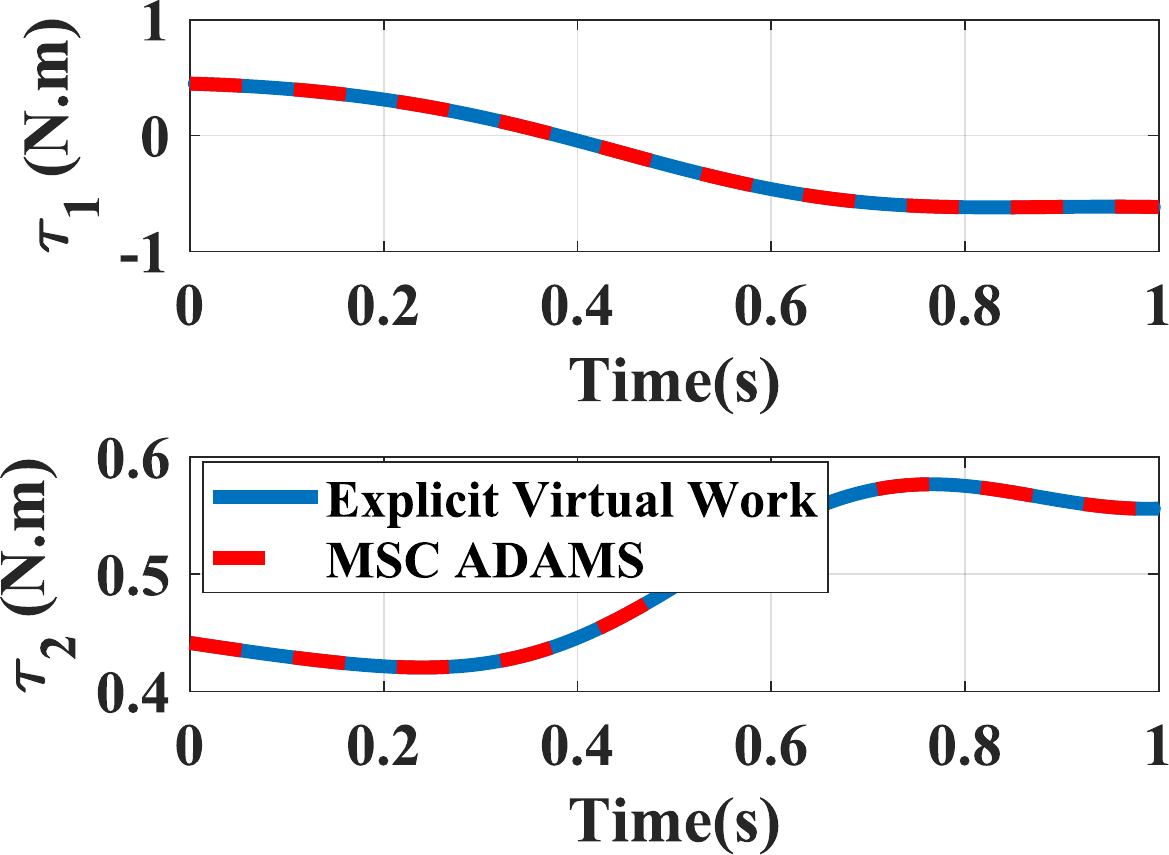}}
\hfil
\subfloat[Error between the actuator torques computed through two models \label{fig:ErrorDynamic}]{\includegraphics[width=2.375in]{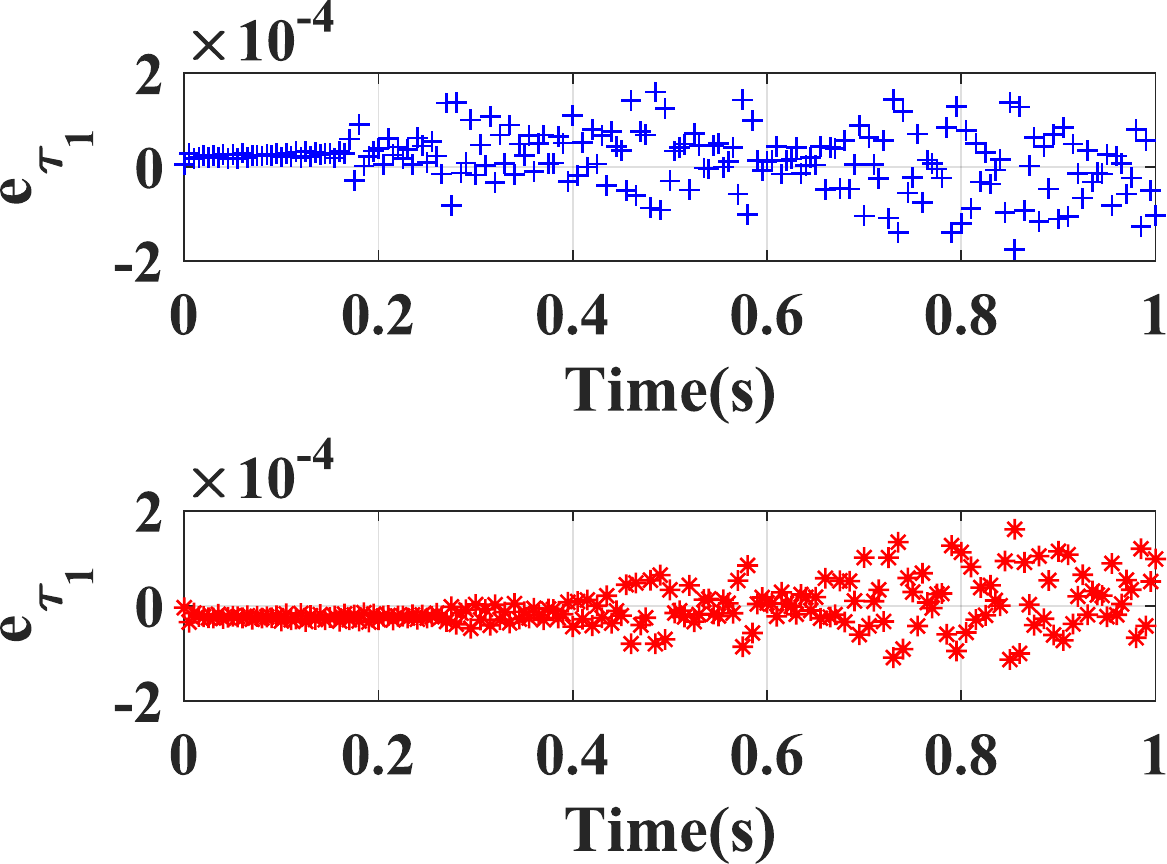}}
\caption{ARAS-Diamond explicit dynamics verification }
\label{fig:Diamnond_Ver}
\end{figure}

In order to verify the resulted dynamic formulation of the
ARAS-Diamond robot, a third-order polynomial trajectory for the
end-effector is considered in the task space, all of which takes
about one second. In the designed trajectory, $\phi$ changes from
$\phi_{0}=0$ to $\phi_{f}=120 ^{\circ}$, while $\gamma$ varies from
$\gamma_0=70^{\circ}$ to $\gamma_f=10 ^{\circ}$.
\textcolor{black}{Furthermore, the \textcolor{black}{considered parameters} of the
ARAS-Diamond model are also reported in
\autoref{Table:Parameters_Diamond}.} By applying this trajectory to
the explicit dynamic formulation of the robot in MATLAB and the
robot's model in MSC-ADAMS\textsuperscript{\textregistered}, the
required actuators torques are determined as shown in
Fig.~\ref{fig:Dynamic}. The difference between the torques generated
by both models is shown in Fig.~\autoref{fig:ErrorDynamic}.

As shown in Figs.~\ref{fig:Dynamic} and \ref{fig:ErrorDynamic}, the
explicit dynamic formulation of the ARAS-Diamond robot is verified
with an accuracy of approximately $10^{-4}~N.m$. This further
confirms the accuracy of the derived kinematic and dynamic
formulations. However, it should be noted that the main reason
caused behind the validation error is the inaccuracy in measurement
and parameter settings in the
MSC-ADAMS\textsuperscript{\textregistered} package. For a more
definitive verification, other trajectories were tested, such as a
sinusoidal and a multi-sine trajectory. In all these cases, as
expected, the resulting error remain in the order of $10^{-4}~N.m$.

\begin{figure}[!t]
\centering
\subfloat[Linear form verification \label{fig:ErrorReg}]{\includegraphics[width=2.3in]{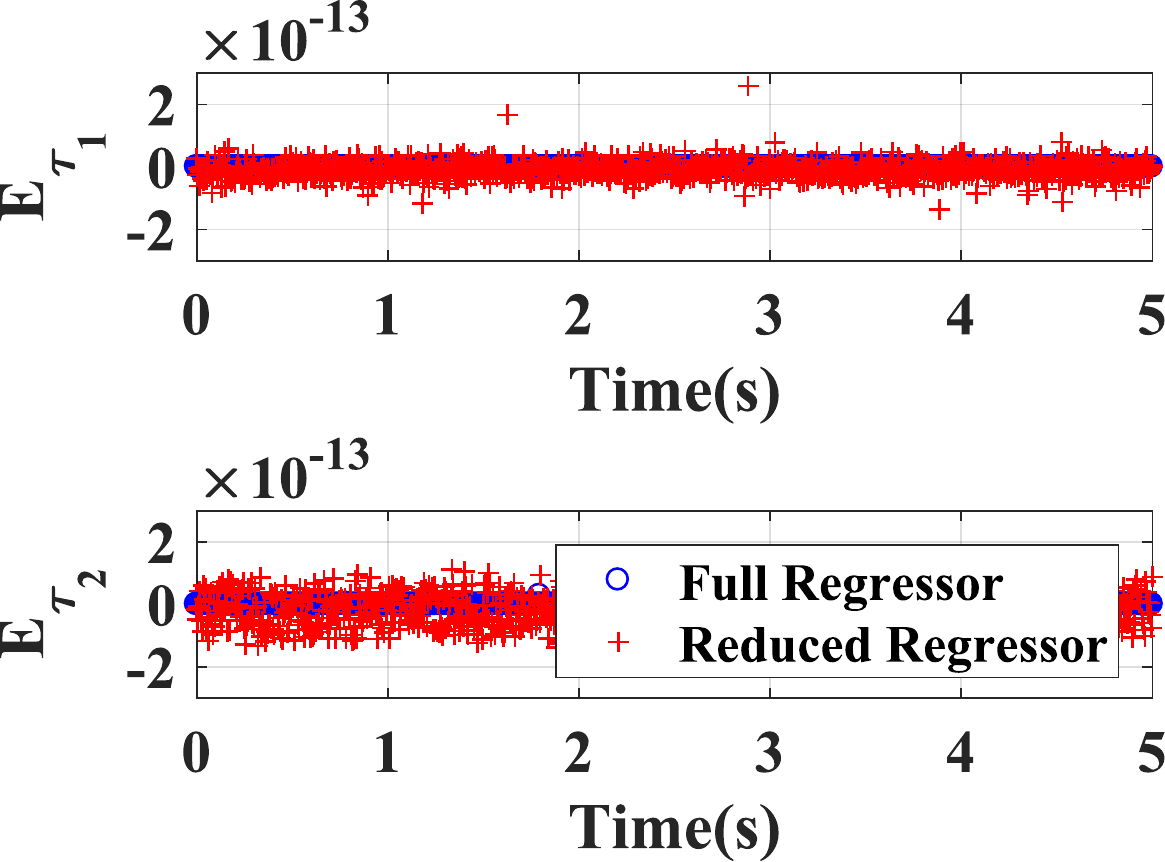}}
\hfil
\subfloat[S-L regressor verification \label{fig:ErrorReg_Slotine}]{\includegraphics[width=2.3in]{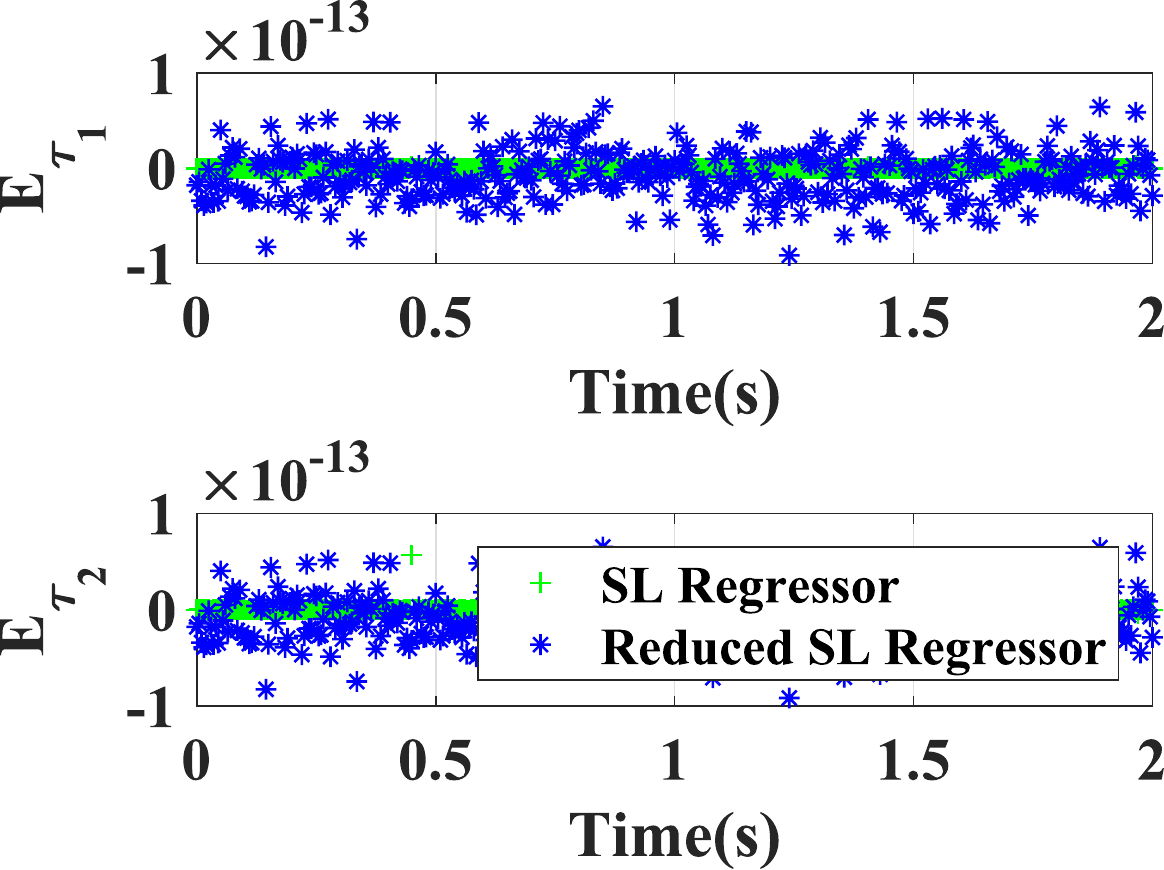}}
\caption{ARAS-Diamond linear regressors verification}
\label{fig:Diamnond_Ver2}
\end{figure}

Additionally, in order to verify the linear regressor form of the
dynamic formulation and the corresponding reduced regressor form, a
random trajectory ($\btheta,\dtheta,\ddtheta$) is generated in the
feasible workspace. Similarly, for the validation of the Slotine-Li
regressor and its corresponding reduced regressor, a random
trajectory for ($\btheta,\dtheta, {\dtheta_r},\ddtheta_r$) is
considered in the feasible workspace of the robot. In this study,
the numerical difference between the \textcolor{black}{left-hand}
sides of the equations of \eqref{eq:Ver_Reg} and
\eqref{eq:Ver_Reg_Slotine}, representing the explicit dynamics of
the robot, and their \textcolor{black}{right-hand sides}, are
calculated and reported in Figs.~\ref{fig:ErrorReg} and
\ref{fig:ErrorReg_Slotine}, respectively.

\textcolor{black}{As it can be seen in these figures, the linear
form of the dynamics, the Slotine-Li regressor, as well as their
corresponding reduced forms differ from the explicit dynamics of the
ARAS-Diamond robot with an order of $10^{-13}$ $N.m$. This provides
another validation of the proposed formulations.}

\subsection{3-\underline{R}RR Spherical Parallel Manipulator Verification}\label{Ver:3RRR}
\begin{figure}[!b]
\centering
\subfloat[Linear form verification \label{fig:Dynamic_3RRR}]{\includegraphics[width=2.3in]{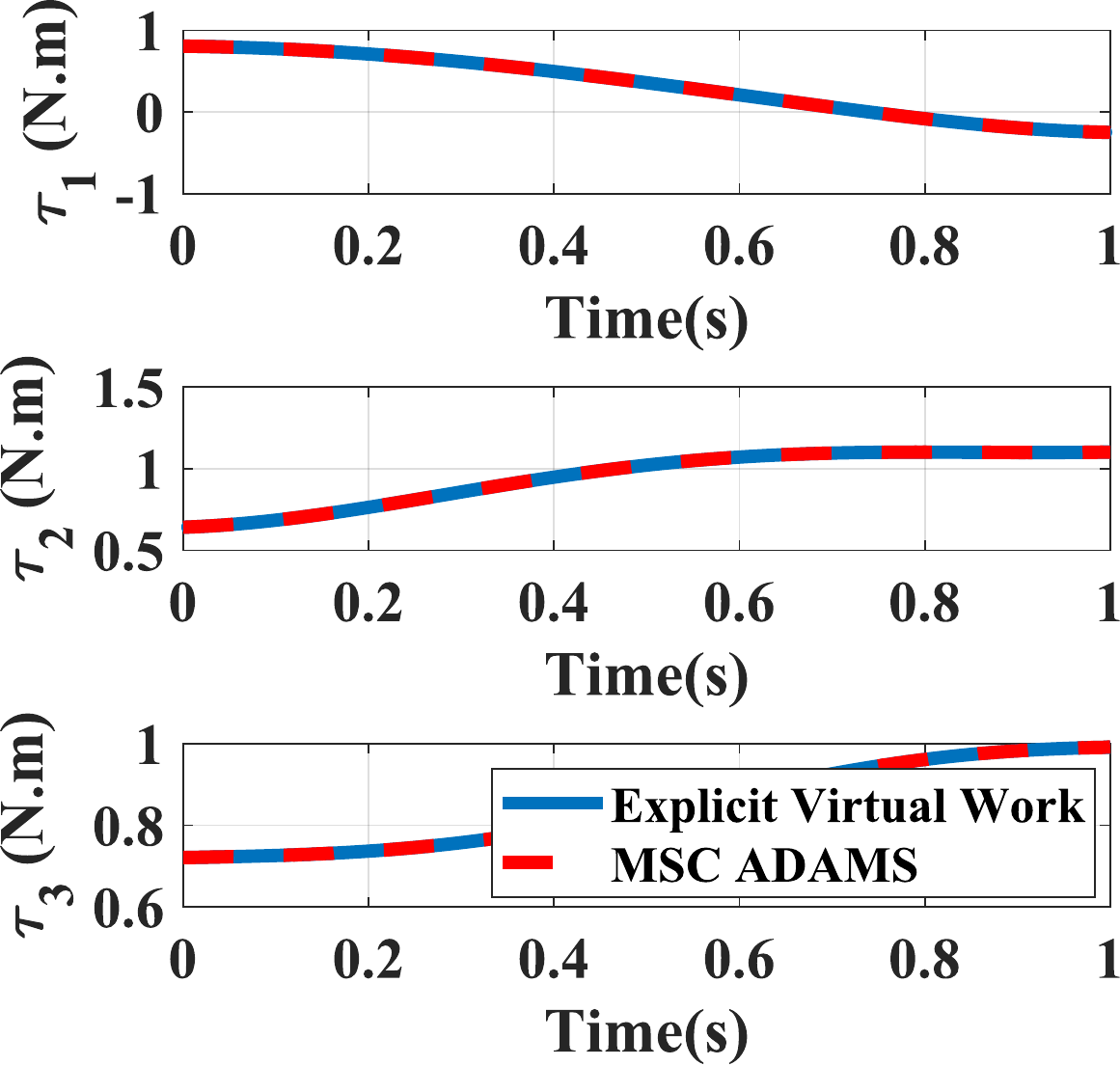}}
\hfil
\subfloat[S-L regressor verification \label{fig:ErrorDynamic_3RRR}]{\includegraphics[width=2.375in]{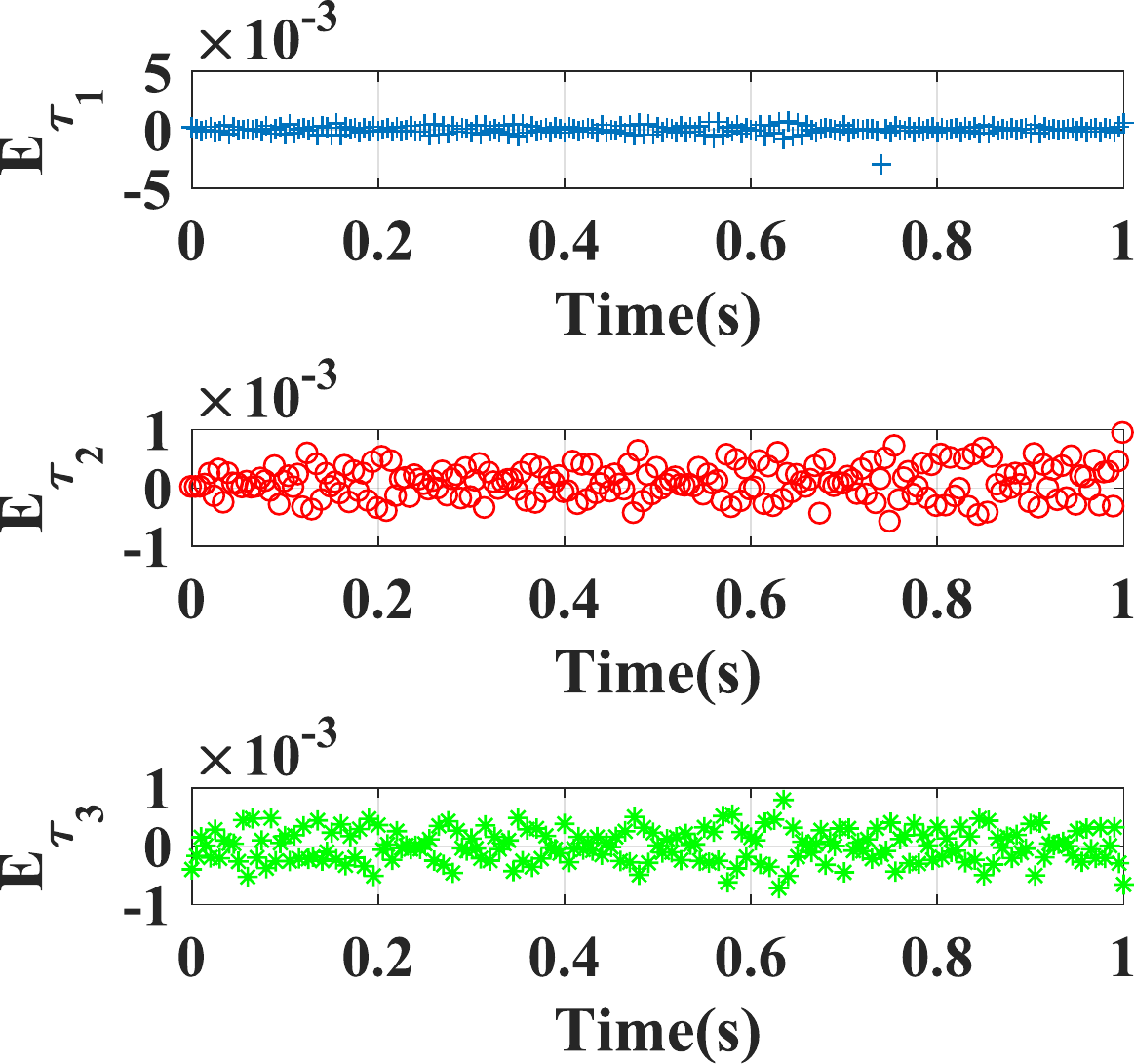}}
\caption{3-\underline{R}RR manipulator explicit dynamics verification}
\label{fig:3RRR_Ver}
\end{figure}

\textcolor{black}{The same method described in section
\ref{Ver:ARAS-Diamond} is adopted here in order to verify the proposed dynamic formulation of the 3-\underline{R}RR spherical
parallel manipulator.} For this purpose, a third-order polynomial trajectory is considered for the moving platform. In this
trajectory, the values of $\btheta$ changes from
$\btheta_0=[0,0,0]^T$ to
$\btheta_f=[10^{\circ},30^{\circ},20^{\circ}]^T$ in one second.
\textcolor{black}{Besides, the \textcolor{black}{parameters} of the considered
3-\underline{R}RR spherical parallel manipulator model have also
been reported in Table \ref{Table:Parameters_3RRR}.} The results of
the dynamic verification and the validation errors are shown in
Figs.~\ref{fig:Dynamic_3RRR} and \ref{fig:ErrorDynamic_3RRR},
respectively.

\textcolor{black}{In addition, the reduced regressors and the
linear and Slotine-Li regressors were compared with the explicit
dynamics of the manipulator over a random trajectory within the
robot feasible workspace.}
\textcolor{black}{To prevent the article from becoming bulky, the error plots are not re-reported. The results show $10^{-14}~N.m$ accuracy between the results, and are available to everyone at the Github page.}


%

\section{Conclusions}

\textcolor{black}{In this paper, different forms of dynamic
formulations of spherical parallel robots (SPRs) including explicit
dynamics, linear and Slotine-Li (S-L) regressors were formulated to
be used in the design of model-based controllers and dynamic
identification schemes.} \textcolor{black}{To this end, the implicit
dynamic of SPRs was first formulated using the principle of virtual
work in task-space, and then by an extension, their explicit dynamic formulation was derived.} \textcolor{black}{The linear and S-L
regressor forms of SPRs were then obtained using explicit dynamic,
and by using the Gauss-Jordan procedure, regressor forms are reduced
to a unique and closed-form structure.} Finally, to verify the
proposed formulations, \textcolor{black}{two SPR case studies},
namely, the ARAS-Diamond robot and the 3-\underline{R}RR spherical
manipulator examined. The results of the explicit dynamics, linear
regression form of robots' dynamics, S-L regressor, and the
corresponding reduced form regressors were verified by comparing
them with those obtained by
MSC-ADAMS\textsuperscript{\textregistered} package. The proposed method opens the possibility of implementing a wide range of
adaptive model-based controllers and regressor-based identification
schemes on SPRs.

\section*{Declaration of Competing Interest}
The authors declare that they have no conflict of interest.

\section*{Acknowledgments}
\textcolor{black}{The authors appreciate the support from Iranian
National Science Foundation (INSF) under grant number 99028112.
Moreover, the authors greatly appreciate Mr. Sina Allahkaram, Mr.
Rohollah Khorambakht, and Mr. Omid Mehdizade for their excellent
feedbacks during this research.}

\appendix

\section{Kinematic Parameters of the ARAS-Diamond}\label{Appendix: Diamond}

The $h$ scalar parameter in equation \eqref{eq:VelJac} is defined as follows:
\begin{equation}\label{eq:K}
h=\frac{\cot \alpha-\cos \hat{A} \cdot \cot \gamma }{\sin \hat{A}}.
\end{equation}

It was stated that equation of $\dot {\hat{B}}=- \dot
{\hat{C}} = s~\dot \gamma$ may be derived in such a way that $s$ is expressed as:
\begin{equation}\label{eq:Q}
s=\frac{\cos \gamma \cdot \sin \hat{A} +h \sin \gamma \cdot \cos \hat{A} }{\cos \hat{B} \cdot \sin \beta}.
\end{equation}

Furthermore, $c$ scalar parameter in the equation of $\dot h= c~\dot \gamma$ may be derived as:
\begin{equation}
c={\frac {h \cdot \sin \gamma \left( \cos \alpha \cdot
\cos \hat{A} -\sin \alpha \cdot \cos \gamma \right)-\sin \hat{A} \cdot \cos \hat{A} \cdot \sin \alpha }{ \sin^{2} \gamma
\cdot \sin \alpha \cdot \sin^{2} \hat{A} }}.
\end{equation}

In order to evaluate the derivative of the Jacobian matrix of the third and
the fourth links in equation of \eqref{eq:jacobian3dot}, it is necessary to compute the derivative of
$s$, which yields to:
\begin{equation}\label{eq:Ali}
\dot s=\frac{{\dot s}_n}{{\dot s}_d},
\end{equation}
in which,
\begin{equation}
\begin{split}
{\dot s}_n= &+h \dot \gamma \cdot \cos \hat B \left(-h \sin \hat A \cdot \sin \gamma + \cos \hat A \cdot \cos \gamma \right)
+s \dot \gamma \cdot \sin \hat B \left(+h \cos \hat A \cdot \sin \gamma + \sin \hat A \cdot \cos \gamma \right) \\
&+ \cos \hat B \left( \dot \gamma \left(+h \cos \gamma \cdot \cos \hat A - \sin \hat A \cdot \sin \gamma \right) + c \dot \gamma \cdot \sin \gamma \cdot \cos \hat A \right ),
\end{split}
\end{equation}

\begin{equation}
{\dot s}_d= {\cos ^2 \hat B} \cdot \sin \beta.
\end{equation}

\section{Kinematic Parameters of the 3-\underline{R}RR Manipulator}\label{Appendix:3RRR_Kinematic}
As mentioned before, the acceleration formulations of the 3-\underline{R}RR
manipulator are based on the derivative of the vectors of $\bp_{1_i}$,
$\bp_{2_i}$, and the scalar parameter of $h_i$.
Using
equations of \eqref{eq:Unit_Vectors_UW} and \eqref{eq:Unit_Vectors_V}, \textcolor{black}{the following formulas are easily extractable:}
\begin{subequations}
\begin{align}
&\dhbu_i=0,\\
&\dhbv_i=~^{0}\mathrm{\dbR_p}~{\hbv_i}^*,\\
&\dhbw_i=\bR_z(\lambda_i)\bR_x(\gamma-\pi)\dbR_z(q_i)\bR_x(\alpha_{1_i})[0,0,1]^T,
\end{align}
\end{subequations}
in which, $\dbR_z$ and $^{0}\mathrm{\dbR_p}$ denote the time
derivative of the rotation matrices of $\bR_z$ and $^{0}\mathrm{\bR}_p$,
respectively. Therefore, according to Eq.~\eqref{eq:Extra_Parameters_3RRR},
the following derivatives may easily be determined:
\begin{subequations}
\begin{align}
&\dbp_{1_i}=\dhbv_i \times \hbw+\hbv \times \dhbw_i,\\
&\dbp_{2_i}=\hbu \times \dhbv_i,\\
&\dot h_{i}= \dbp_{1_i} \cdot \hbu.
\end{align}
\end{subequations}

\section{Parameters for Verification}
\begin{table}[ht!]
\caption{Considered parameters of ARAS-Diamond \label{Table:Parameters_Diamond}}
\centering
\begin{tabular}{@{\extracolsep{\fill}}c c c}
\hline
Symbol& Description&Value\\
\hline
$\alpha=\beta$& \textcolor{black}{Angular length of links} & $45^{\circ}$ deg \\
${m}$& Mass of links & $[0.117,0.112,0.155,0.145]^T$ kg \\
$^{1}\mathrm{\brho}_{1}$ & CM of link1 & $[0.098,0,0.232]^T$m \\
$^{2}\mathrm{\brho}_{2}$ & CM of link2 & $0.087,0,0.210]^T$m \\
$^{3}\mathrm{\brho}_{3}$ & CM of link3 & $[ 0.107,0,0.254]^T$m \\
$^{4}\mathrm{\brho}_{4}$ & CM of link4 & $[0.078,0,0.188]^T$m \\
$^{1}\mathrm{\bI}_{1}$ & MI of link1 & $diag(6.440,6.350,1.716)10^{-4}$ kg.$m^2$ \\
$^{2}\mathrm{\bI}_{2}$ & MI of link2 & $diag(5.359,5.284,0.150)10^{-4}$ kg.$m^2$ \\
$^{3}\mathrm{\bI}_{3}$ & MI of link3 & $diag(9.849,9.342,0.610)10^{-4}$ kg.$m^2$ \\
$^{4}\mathrm{\bI}_{4}$ & MI of link4 & $diag(7.577,7.496,2.348)10^{-4}$ kg.$m^2$ \\
\hline
\end{tabular}
\end{table}



\begin{table}[ht!]
\caption{Considered parameters of the considered 3-\underline{R}RR SPR \label{Table:Parameters_3RRR}}
\centering
\begin{tabular}{@{\extracolsep{\fill}}c c c}
\hline
Symbol& Description&Value\\
\hline
$\etab=\blambda$& \textcolor{black}{Structural angles} & $[0,120^{\circ},240^{\circ}]^T$ deg \\
$\beta$& \textcolor{black}{Base regular pyramid} & $\arccos(\frac{\sqrt{3}}{3})$ deg \\
$\gamma$& \textcolor{black}{Moving platform regular pyramid} & $\arcsin(\frac{\sqrt{3}}{3})$ deg \\
$\alpha_{1_i}$& \textcolor{black}{Proximal links length} & $80^{\circ}$ deg \\
$\alpha_{2_i}$& \textcolor{black}{Distal links length} & $\alpha_{2_i}=70^{\circ}$ deg \\
$m_p$& Mass of moving platform & $0.604$ kg \\
$m_{1_1}=m_{1_2}=m_{1_3}$& Mass of proximal links& $0.501$ kg \\
$m_{2_1}=m_{2_2}=m_{2_3}$& Mass of distal links & $0.389$ kg \\
$^{p}\mathrm{\brho}_{p}$& CM of moving platform & $\left[0,0,0.084\right]^T$ m \\
$^{1}\mathrm{\brho}_{1_1}=~^{1}\mathrm{\brho}_{1_2}=~^{1}\mathrm{\brho}_{1_3}$& CM of proximal links & $\left[0,-0.117,0.139\right]^T$ m \\
$^{2}\mathrm{\brho}_{2_1}=~^{2}\mathrm{\brho}_{2_2}=~^{2}\mathrm{\brho}_{2_3}$& CM of distal links & $\left[0,-0.093,0.133\right]^T$ m \\
$^{p}\mathrm{\bI}_{p}$& MI of moving platform & $diag(0.003,0.001,0.001)$ kg.$m^2$\\
$^{1}\mathrm{\bI_{1_1}}=~^{1}\mathrm{\bI_{1_2}}=~^{1}\mathrm{\bI_{1_3}}$& MI of proximal links & $diag(0.003,0.003,0)$ kg.$m^2$ \\
$^{2}\mathrm{\bI_{2_1}}=~^{2}\mathrm{\bI_{2_2}}=~^{2}\mathrm{\bI_{2_3}}$& MI of distal links & $diag(0.001,0.001,0)$ kg.$m^2$\\
\hline
\end{tabular}
\end{table}

\section*{References}
\bibliography{sample.bib}

\end{document}